# Evaluation of Thermal Imaging on Embedded GPU Platforms for Application in Vehicular Assistance Systems

Muhammad Ali Farooq, Waseem Shariff, Peter Corcoran, Fellow, IEEE

***Abstract*—This study is focused on evaluating the real-time performance of thermal object detection for smart and safe vehicular systems by deploying the trained networks on GPU & single-board EDGE-GPU computing platforms for onboard automotive sensor suite testing. A novel large-scale thermal dataset comprising of > 35,000 distinct frames is acquired, processed, and open-sourced in challenging weather and environmental scenarios. The dataset is a recorded from lost-cost yet effective uncooled LWIR thermal camera, mounted stand-alone and on an electric vehicle to minimize mechanical vibrations. State-of-the-art YOLO-V5 networks variants are trained using four different public datasets as well newly acquired local dataset for optimal generalization of DNN by employing SGD optimizer. The effectiveness of trained networks is validated on extensive test data using various quantitative metrics which include precision, recall curve, mean average precision, and frames per second. The smaller network variant of YOLO is further optimized using TensorRT inference accelerator to explicitly boost the frames per second rate. Optimized network engine increases the frames per second rate by 3.5 times when testing on low power edge devices thus achieving 11 fps on Nvidia Jetson Nano and 60 fps on Nvidia Xavier NX development boards.**

## I. Introduction

Thermal imaging is the digital interpretation of the infrared radiations emitted from the object. Thermal imaging cameras with microbolometer focal plane arrays (FPA) is a type of uncooled detector that provides low-cost solutions for acquiring thermal images in different weather and environmental conditions. These cameras when integrated with AI-based imaging pipelines can be used for various real-world applications. In this work, the core focus is to design an intelligent thermal object detection-based video analysis system for automotive sensor suite application that should be effective in all light conditions thus enabling safe and more reliable road journeys. Unlike other video solutions such as visible imaging which mainly relies on reflected light thus having the greater chances of being blocked by visual impediments, thermal imaging does not require any external lighting conditions to capture quality images and it can see through visual obscurants such as dust, light fog, smoke, or other such occlusions. Moreover, the integration of AI-based thermal imaging systems can provide us with a multitude of advantages from better analytics with fewer false alarms to increased coverage, provide redundancy and, higher return on investment.

In this research work, we have focused on utilizing thermal data for designing efficient AI-based object detection and classification pipeline for Advance Driver-Assistance Systems. Such type of thermal imaging-based forward sensing (F-sense) system is useful in providing enhanced safety and security features thus enabling the driver to better scrutinize the complete road-side environment. For this purpose, we have used a state-of-the-art (SoA) end-to-end deep learning framework YOLO-V5 on thermal data. In the first phase, a novel thermal dataset is acquired for training and validation purposes of different network variants of YOLO-V5. The data is captured using a prototype low-cost uncooled LWIR thermal camera specifically designed under the ECSEL Heliaus research project [32]. The raw thermal data is processed using shutterless camera calibration, automatic gain control, bad-pixel removal, and temporal denoising methods.

Furthermore, the trained network variants are deployed and tested on two state-of-the-art embedded GPU platforms, which include NVIDIA Jetson nano [23] and Nvidia Jetson Xavier NX [25]. Thus, studying the extensive real-time and on-board feasibility in terms of various quantitative metrics, inference time, FPS, and hardware sensor temperatures.

The core contributions of the proposed research work are summarized below:

- Preparation and annotation of a large open-access dataset of thermal images captured in different weather and environmental conditions.
- A detailed comparative evaluation of SoA object detection based on a modified YOLO-V5 network, fine-tuned for thermal images using this newly acquired dataset.
- Model optimization using TensorRT inference accelerator to implement a fast inference network on SoA embedded GPU boards (Jetson, Xavier) with comparative evaluations.

October 25th, 2021, "This research work is funded by the ECSEL Joint Undertaking (JU) under grant agreement No 826131 (Heliaus project https://www.heliaus.eu/ ). The JU receives support from the European Union's Horizon 2020 research and innovation program and National funders from France, Germany, Ireland (Enterprise Ireland, International Research Fund), and Italy".

Muhammad Ali Farooq, Peter Corcoran, and Waseem Shariff are with the National University of Ireland Galway, (NUIG), College of Science & Engineering Galway, H91TK33, Ireland (e-mail: m.farooq3@nuigalway.ie, peter.corcoran@nuigalway.ie, waseem.shariff@nuigalway.ie).

Thermal Dataset Link: https://bit.ly/3tAkJ0J

GitHub Link: https://github.com/MAli-Farooq/Thermal-YOLO

- A determination of realistic frame rates that can be achieved for thermal object detection on SoA embedded GPU platforms.

## II. Background

ADAS (Advanced Driver Assistance Systems) are classified as AI-based intelligent systems integrated with core vehicular systems to assist the driver by providing a wide range of digital features for safe and reliable road journeys. Such type of system is designed by employing an array of electronic sensors and optical mixtures such as different types of cameras to identify surrounding impediments, driver faults, and reacts automatically.

The second part of this section will mainly summarize the existing/ published thermal datasets along with their respective attributes. These datasets can be effectively used for training and testing the machine learning algorithms for object detection in thermal spectrum for ADAS. The complete dataset details are provided in Table I.

TABLE I
EXISTING THERMAL DATASETS

| Datasets | Condition Day | Condition Night | Annotations | Objects | Total no. of frames | Image Resolution |
|---|---|---|---|---|---|---|
| OSU Thermal [2] | ✓ | ✓ | - | Person, Cars, Poles | 284 | 360 X 240 |
| CVC [19] | ✓ | ✓ | - | Person, Cars, Poles, Bicycle, Bus, Bikes | 11K | 640 X 480 |
| LITIV [3] | - | - | - | Person | 6K | 320 X 240 |
| TIV [4] | - | - | - | Person, Cars, Bicycle, Bat | 63K | 1024 X 1024 |
| SCUT [6] | - | ✓ | ✓ | Person | 211K | 384 X 288 |
| FLIR [7] | ✓ | ✓ | ✓ | Person, Cars, Poles, Bicycle, Bus, Dog | 14K | 640 X 512 |
| KAIST [5] | ✓ | ✓ | ✓ | Person, Cars, Poles, Bicycle, Bus | 95K | 640 X 480 |

### *A. Related Literature*

We can find numerous studies regarding the implementation of object detection algorithms using AI based conventional machine learning as well as deep learning algorithms. Such type of optical imaging-based systems system can be deployed and effectively used as forward sensing methods for ADAS. Advanced Driver-Assistance Systems (ADAS) is an active area of research that seeks to make road trips more safe and secure. Real time object detection plays a critical role to warn the driver thus allowing them to make timely decisions [8]. Ziyatdinov et al [8] proposed an automated system to detect road signs. This method uses the GTSRB dataset [20] to train on conventional machine learning algorithms which include SVM, KNN, and Decision Trees classifier. The results proved that SVM and K – nearest neighbour (k-NN) outperforms all other classifiers. Autonomous cars on the road require the abilities to consistently perceive and comprehend their surroundings [9]. Oliver et al [9] presented a procedure to use Bernoulli particle filter, which is suitable for object identification because it can handle a wide range of sensor measurements as well as object appearance-disappearance. Gang Yan et al [10] proposed a novel method to use HOG to extract features and AdaBoost and SVM classifiers to detect vehicles in real-time. The histogram of oriented gradients (HOG) is a feature extraction technique used for object detection in the domain of computer vision and machine learning. The study concluded that the AdaBoost classification technique performed slightly better than SVM since it uses the ensemble method. Authors in [11], proposed another approach to detect vehicles on road using HOG filters to again extract features from the frames and then classify them using support vector machines and decision tree classification algorithms. Furthermore, SVM achieved 93.75% accuracy, which outperformed decision tree accuracy on classifying the vehicles. These are some of the conventional machine learning object detection techniques used for driver assistance system till date. The main drawback of traditional machine learning techniques is that the features are extracted and predefined prior to training and testing of the algorithms. When dealing with high-dimensional data, and with many classes conventional machine learning techniques are often ineffective [21].

Deep learning approaches have emerged as more reliable and effective solutions than these classic approaches. There are many state-of-the-art pre-trained deep learning classifiers and object detection models which can be retrained and rapidly deployed for designing efficient forward sensing algorithms [22]. YOLO (you only look once) object classifier provides sufficient performance to operate at real-time speeds on conventional video data without compromising the overall detector precision [15]. Veta et al [12] presented a technique for detecting objects at a distance by employing YOLO on low-quality thermal images. Another research [13] focused on pedestrian detection in thermal images using the histogram of gradient (HOG) and YOLO methods on FLIR [7] dataset and computed performance with a 70% accuracy on test data using the intersection over union technique. Further, Rumi et al [14] proposed a real-time human detection technique using YOLO-v3 on KAIST [5] thermal dataset, achieving 95.5% average precision on test data. Authors in [16] proposed a human detection system using YOLO object detector. The authors used their custom dataset recorded in different weather conditions using FLIR Therma-CAM P10 thermal camera.

Focusing on road-side objects, authors in [17] used YOLO-v2 object detection model to enhance the recognition of tiny vehicle objects by combining low-level and high-level features of the image. In [18], the authors proposed a deep learning-based vehicle occupancy detection system in a parking lot using a thermal camera. In this study authors had established that YOLO, Yolo-Conv, GoogleNet, and ResNet18 are computationally more efficient, take less processing time, and are suitable for real-time object detection. In one of the most recent studies [24], the efficacy of typical state-of-the-art object detectors which includes Faster R-CNN, SSD, Cascade R-CNN, and YOLO-v3 was assessed by retraining them on a thermal dataset. The results demonstrated that Yolo-v3 outclassed other object SoA object detectors.

### *B. Object Detection on Edge Devices*

AI on edge devices benefits us in various methods such that it speeds up decision-making, makes data processing more reliable, enhances user experience with hyper-personalization, and cuts down the costs. While machine learning models have shown immense strength in diversified consumer electronic applications, the increased prevalence of AI on edge has contributed to the growth of special-purpose embedded boards for various applications. Such type of embedded boards can achieve AI inference at higher frames per second (fps) and low power usage. Some of these board includes Nvidia Jetson Nano, Nvidia Xavier, Google Coral, AWS DeepLens, and Intel AI-Stick. Authors in [26-27] proposed a raspberry pi-based edge computing system to detect thermal objects. Sen Cao et al [28] developed a roadside object detector using KITTI dataset [29] by training an efficient and lightweight neural network on Nvidia Jetson TX2 embedded GPU [28].

In another study [30] authors proposed deep learning-based smart task scheduling for self-driving vehicles. This task management module was implemented on multicore SoCs (Odroid Xu4 and Nvidia Jetson).

The overall goal of this study is to analyse the real-time performance feasibility of Thermal-YOLO object detector by deploying on edge devices. Different network variants of yolo-v5 framework are trained and fine-tuned on thermal image data and implemented on the Nvidia Jetson Nano [23] and Nvidia Jetson Xavier NX [25]. These two platforms, although from the same manufacturer provide very different levels of performance and may be regarded as close to current SoA in terms of performance for embedded neural inference algorithms.

## III. Thermal Data Acquisition at Scale for ADAS

This section will mainly cover the thermal data collection process using the LWIR prototype thermal imaging camera. The overall data is consisting of more than 35K distinct thermal frames acquired in different weather and environmental conditions. The data collection process includes shutterless camera calibration and thermal data processing [36], using the Lynred Display Kit (LDK) [1], data collection methods, and overall dataset attributes with different weather and environmental conditions for comprehensive data formation.

### *A. Prototype Thermal Camera*

For the proposed research work we have utilized micro-bolometer technology based uncooled thermal imaging camera developed under the HELIAUS project [32]. The main characteristic of this camera includes low-cost, lightweight and its sleek compact design thus allowing to easily integrate it with artificially intelligent imaging pipelines for building effective in-cabin driver-passenger monitoring and road monitoring systems for ADAS. It enables us to capture high-quality thermal frames with low-power consumption thus proving the agility of configurations and data processing algorithms in real-time. Fig. 1 shows the prototype thermal camera. The technical specifications of the camera are as follows, the camera type is a QVGA long-wave infrared (LWIR) with a spectral range from 8-14 μm and a camera resolution of 640 X 480 pixels. The focal length (f) of the camera is 7.5 mm, F-number is 1.2, the pixel pitch is 17 μm, and the power consumption is less than 950mW. The camera relates to a high-speed USB 3.0 (micro-USB) port for the interface.

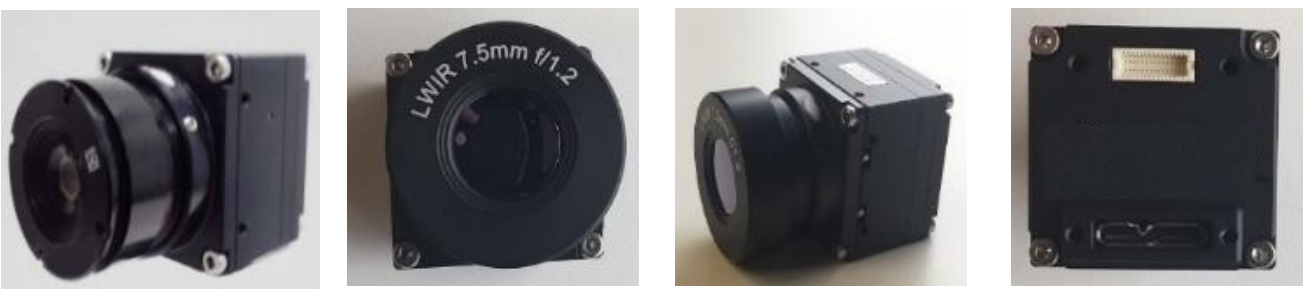

**Fig. 1**. LWIR thermal imaging module images from different view angles.

The data is recorded using a specifically designed toolbox. The complete camera calibration process along with the data processing pipeline is explained in the next section.

### *B. Shutterless Calibration and Real-time Data Processing*

This section will highlight the thermal camera calibration process for shutterless camera configuration along with real-time data processing methods for converting the raw thermal data to refined outputs. Shutterless technology allows uncooled IR engines and thermal imaging sensors to continuously operate without the need for a mechanical shutter for Non-Uniformity Correction (NUC) operations. Such type of technology provides proven and effective results in poor visibility conditions ensuring good quality thermal frames in real-time testing situations. For this, we have used a low-cost blackbody source to provide three different constant reference temperature values referred to as T-ambient1-BB1 (hot uniform scene with temperature value of 40 degree centigrade), T-ambient1-BB2 (cold uniform scene with the temperature value of 20 degree centigrade), and T-ambient2-BB1 (either hot or cold uniform scene but with different temperature value). The imager can store up to 50 snapshots and select the best uniform temperature scenes for calibration purposes. Fig. 2 shows the blackbody used for the thermal camera calibration.

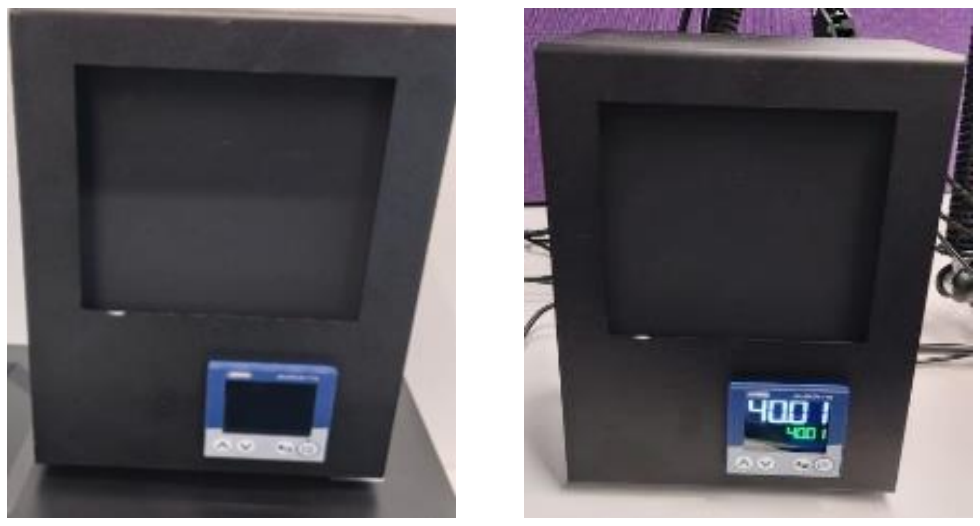

(a) (b)

**Fig. 2.** Thermal camera calibration a) blackbody source used for LWIR thermal camera calibration, b) uniform scene: temperature set to 40.01 degree centigrade.

Once the uniform temperature images are recorded the images are loaded in camera SDK as shown in Fig. 3 to finally calibrate the shutterless camera stream. Fig. 4 shows the results before applying shutterless calibration and processed results using shutterless algorithms on thermal frame capture through the prototype thermal IR camera.

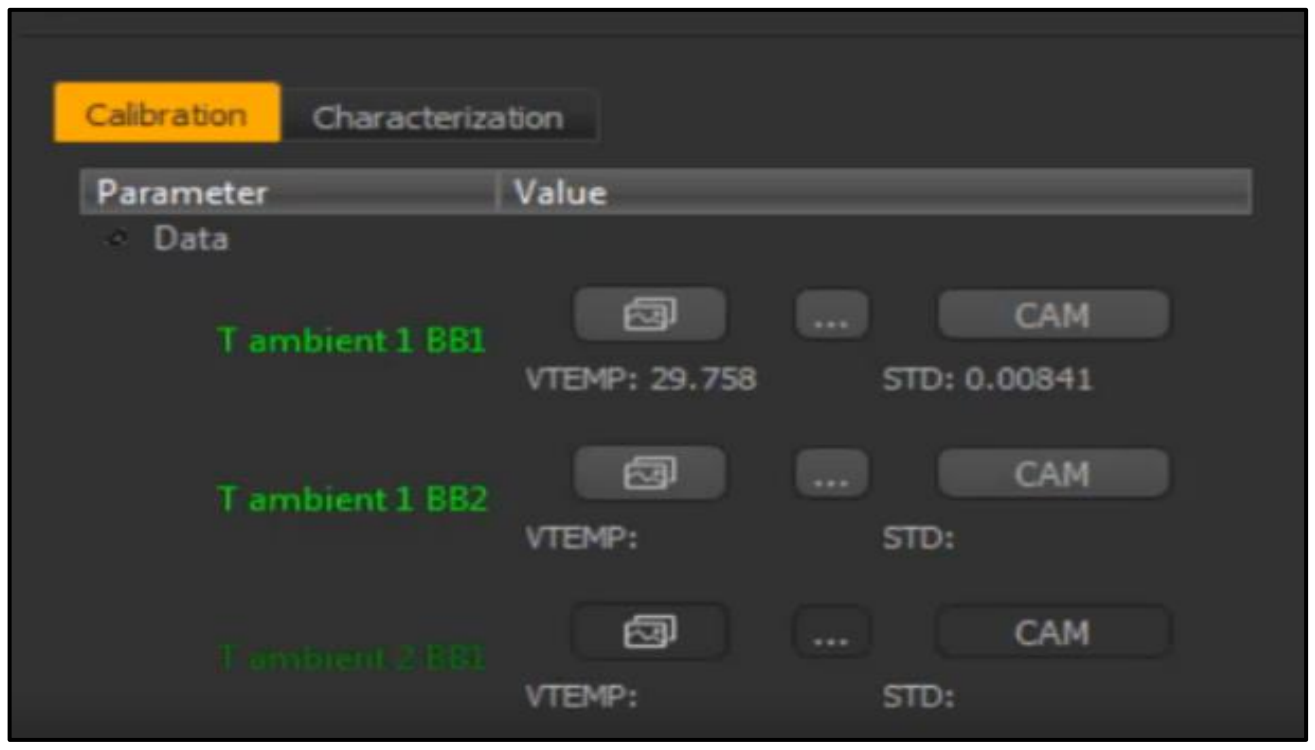

**Fig. 3.** Prototype thermal camera SDK for loading constant reference temperatures values for shutterless camera calibration.

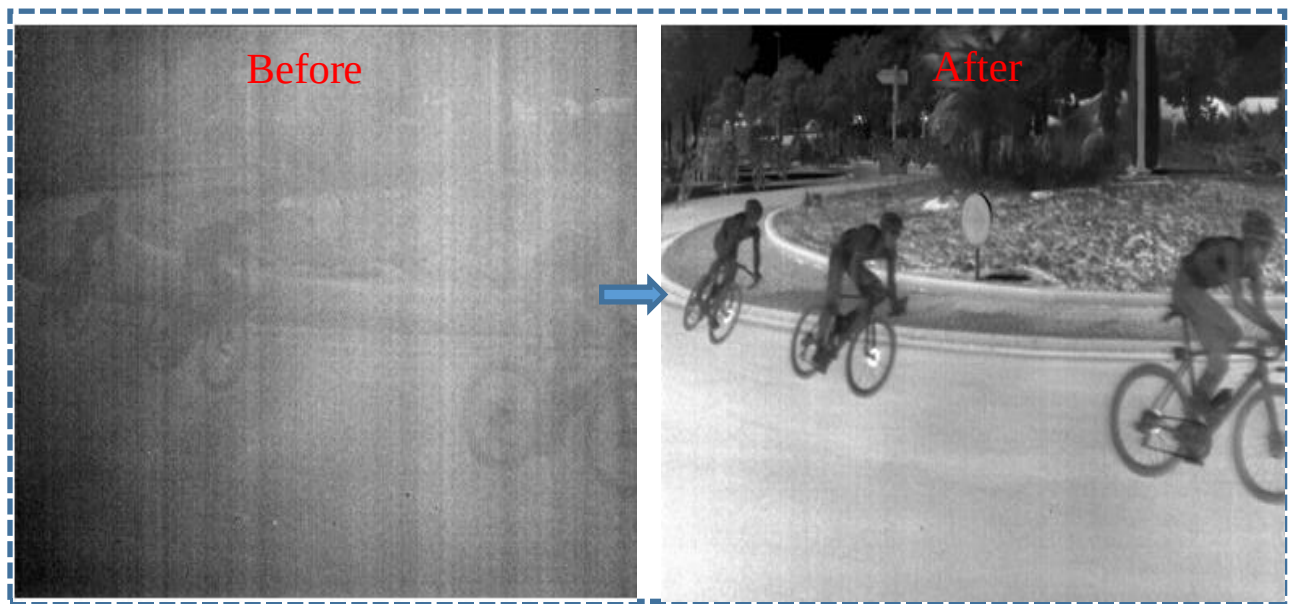

**Fig. 4.** Shutterless algorithm results on sample thermal frame captured from 640x480 LWIR thermal camera designed by Lynred France [1].

In the next phase, various real-time image processing-based correction methods are applied to convert the original thermal data to produce good-quality thermal frames. Fig. 5 shows the complete image processing pipeline.

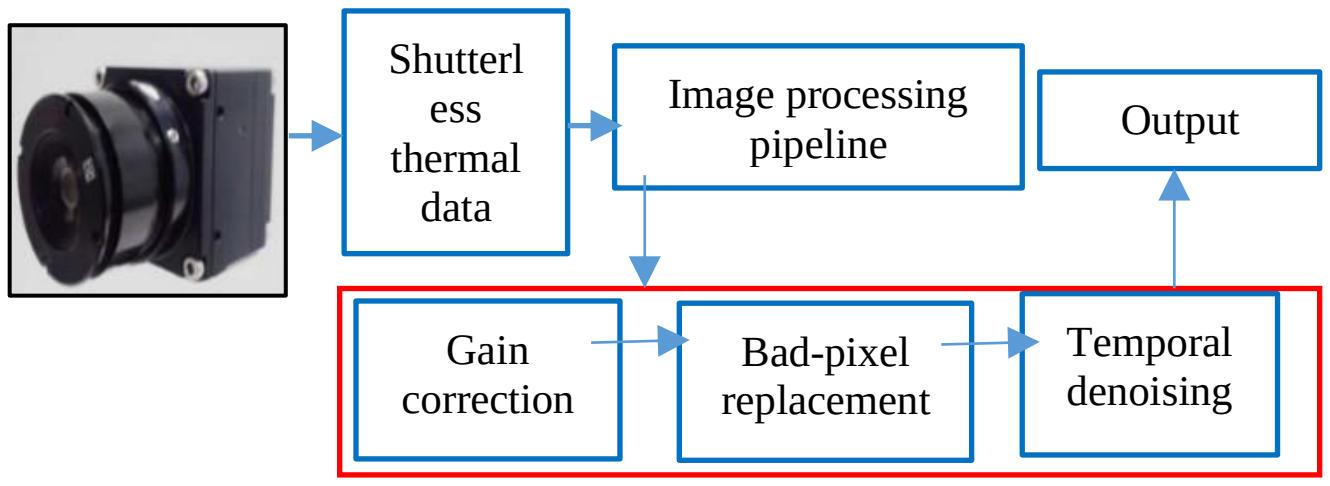

**Fig. 5.** Thermal image correction pipeline

As shown in Fig. 5 image processing pipeline consist of three different image correction methods which include gain correction, bad-pixel replacement, and temporal denoising. The further details of these methods are provided as follows.

1) **Gain Correction Automatic Gain Control (AGC)**
Thermal image detectors, based on flat panels, suffer from irregular gains due to the non-uniform amplifiers. To correct the irregular gains, a common yet effective technique referred to as automatic gain control is applied. It is usually based on the gain map. By averaging uniformly illuminated images without any objects, the gain map is designed. By increasing the number of images for averaging provides a good gain-correction performance since the remained quantum noise in the gain map is reduced [1].

2) **Bad Pixel Replacement (BPR)**
This is used to list bad pixels estimated at the calibration stage. It works by tracking potential new bad pixels by looking at pixel neighbourhood also known as the nearest neighbour method. Once it traces the bad pixels in the nearest neighbour it replaces them with good pixels. Fig. 6 demonstrates one such example.

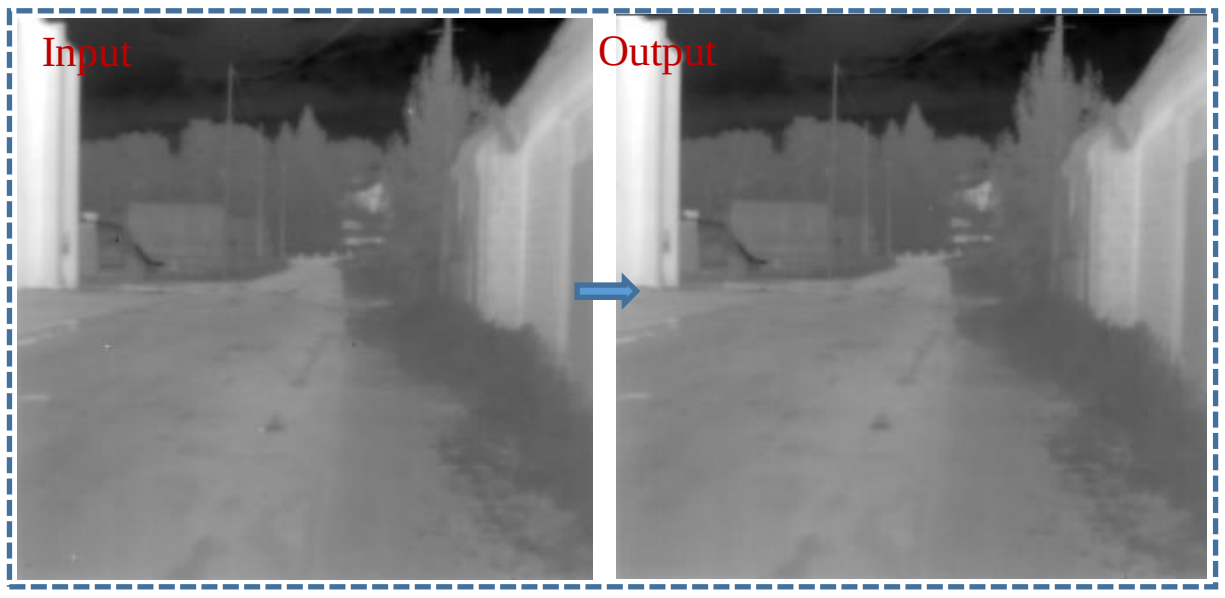

**Fig. 6.** Bad pixel replacement algorithm output on sample thermal frame, left side frame with some bad pixels and the right side is processed frame.

3) **Temporal Denoising (TD)**
The consistent reduction of image noise poses a frequently recurring problem in digitized thermal imaging systems and especially when it comes to un-cooled thermal imagers [34]. To mitigate these limitations for better outputs different methods are used which include hardware as well software-based image processing methods such as temporal and spatial denoising algorithms. The temporal denoising method is used to decrease the temporal noise between different frames of the video. In commercial solutions, it usually works by gathering multiple frames and averaging those frames to cancel out the random noise among the frames. In our data acquisition process, this method is used after applying the shutterless algorithm. Fig. 7 shows the sample thermal images in the form of outcomes after applying shutterless algorithms and all the image processing-based corrections methods as shown in Fig. 5.

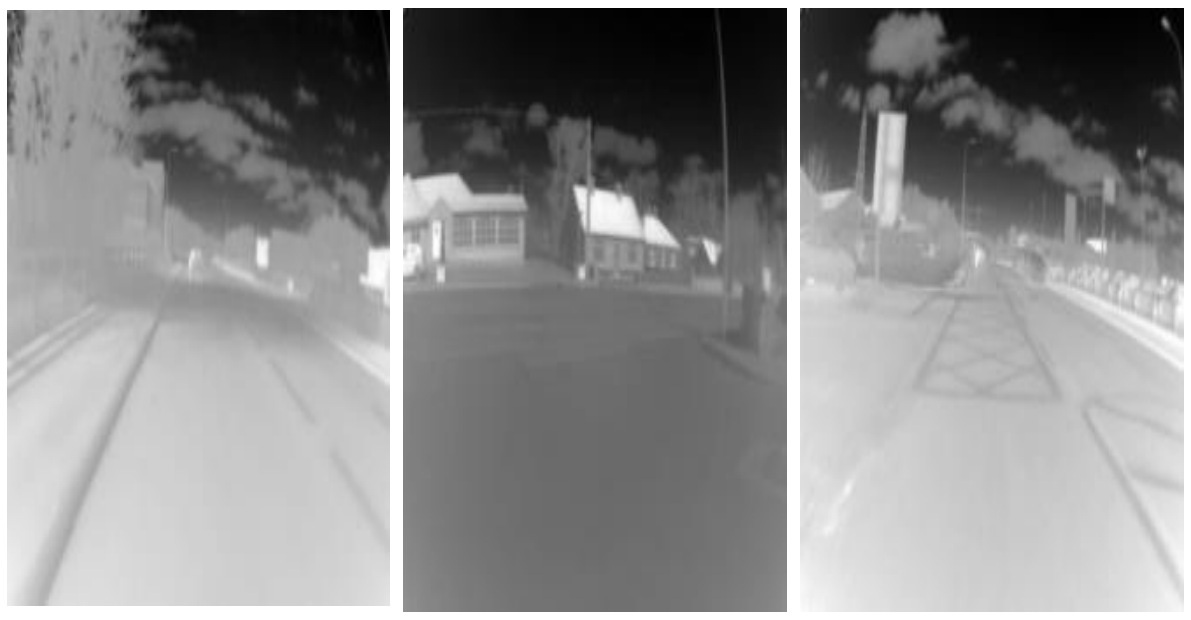

**Fig. 7.** High-quality thermal frames after applying the shutterless calibration algorithm and image correction methods.

### C. Data Collection Methods and Overall Dataset Attributes

This section will highlight different data collection approaches adopted in this research work. The data is collected in two different approaches. In, the first approach (M-1) the data is gathered in an immobile method by placing the camera at a fixed place. The camera is mounted on the tripod stand at a fixed height of nearly 30 inches such that the roadsides objects are covered in the video stream. The thermal video stream is recorded at 30 frames per second (FPS). The data is recorded in different weather and environmental conditions. Fig. 8 shows the M-1 data acquisition setup. In the second method (M-2) the thermal imaging system is mounted over the car and data is acquired in the mobile method. The prime reason for collecting the data in two different methods is to bring variations and collect distinctive local data in different environmental and weather conditions. For this, a specialized waterproof camera housing case was designed to hold the thermal camera in the correct position and angle to cover the entire roadside scene. The housing case is fixed on a suction-based tripod stand thus allowing us to easily fix and remove the complete structure from the car bonnet. The housing case also contains a visible camera to get initial visible images as reference data thus allowing us to adjust both the camera positions in proper angle and field of view.

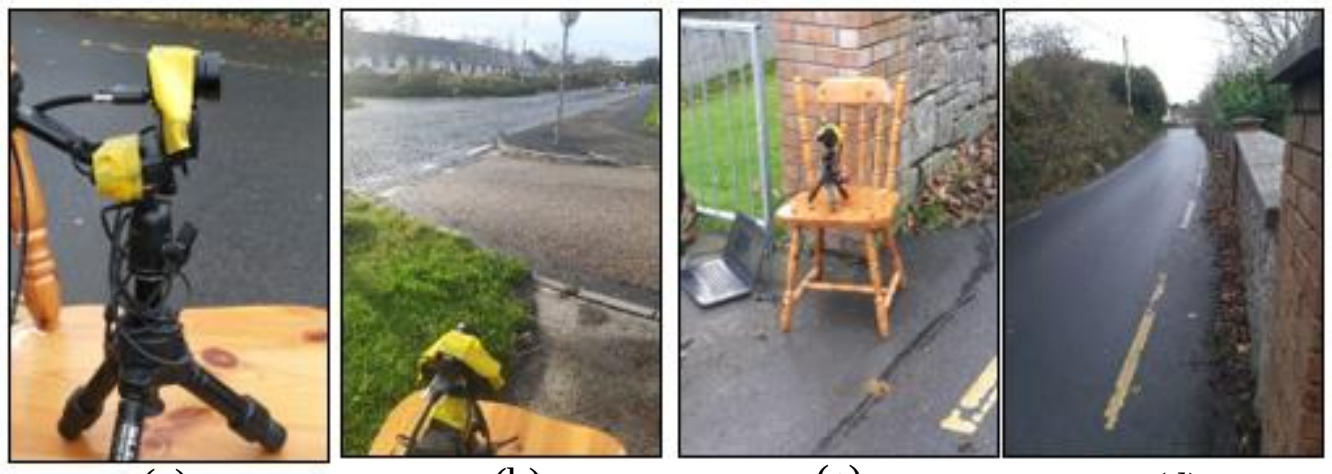

(a) (b) (c) (d)

**Fig. 8.** Data Acquisition setup by placing the camera at a fixed place a) camera mounted on a tripod stand, b) complete daytime roadside view, c) video recording setup at 30fps, d) evening time alleyway view.

Fig. 9 shows the camera housing case along with the initial data acquisition setup whereas

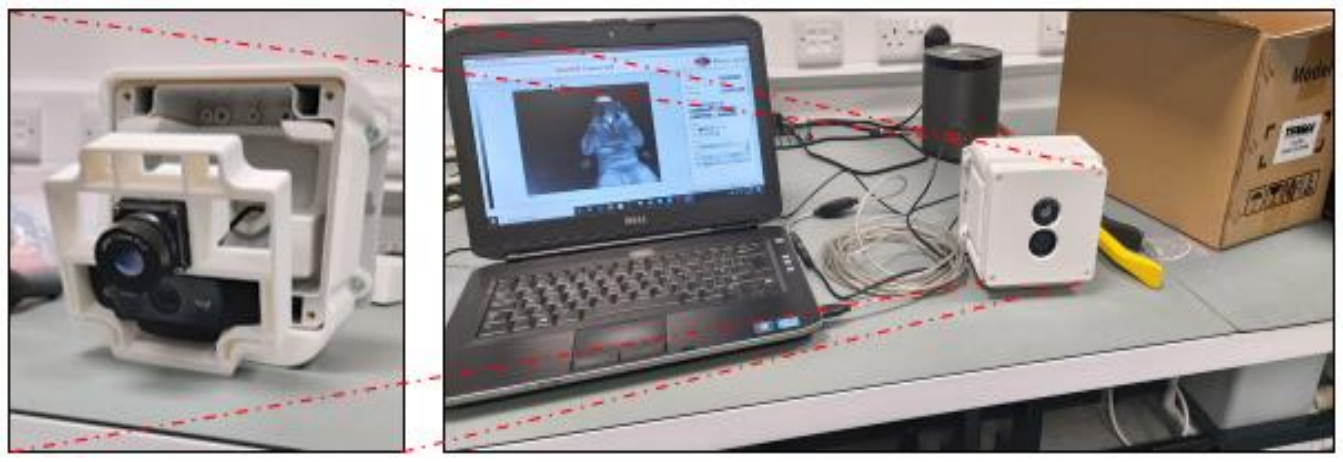

(a) (b)

**Fig. 9.** Data acquisition setup through car a) camera housing case holding thermal and visible camera, b) initial data acquisition testing phase.

Fig. 10 shows the housing case fixed on tripod structure and complete M-2 acquisition setup mounted on the car. The overall dataset is acquired from Galway County Ireland. The data is collected in form of short video clips and more than> 35,000 unique thermal frames have been extracted from the recorded video clips. The data is recorded in the daytime, evening time, and night-time which is distributed in the ratio of 44.61%, 31.78%, and 23.61% respectively of overall data. The complete dataset attributes are summarized in Table II. The acquired data comprises distinct stationary classes, such as road signs and poles, as well as moving object classes such as pedestrians, cars, buses, bikes, and bicycles.

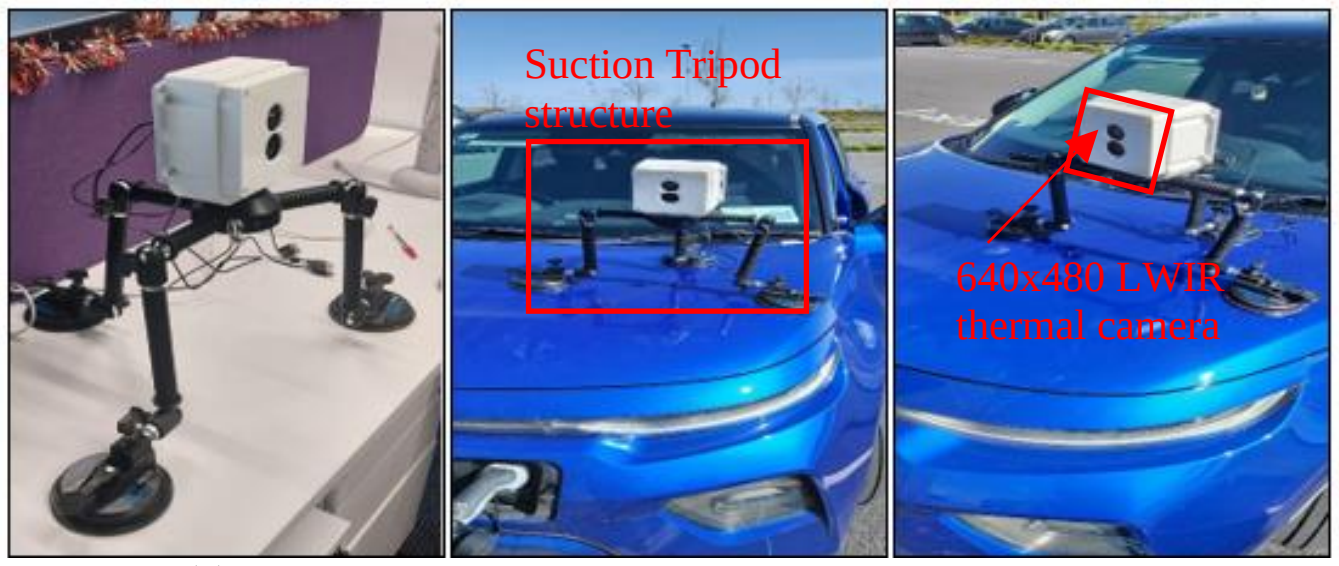

(a) (b) (c)

**Fig. 10.** Complete data acquisition setup mounted on the car a) camera housing fixed on a suction tripod stand, b) data acquisition kit from the front view, c) data acquisition kit from the side view.

TABLE II

NEW THERMAL DATASET ATTRIBUTES

| Locally acquired dataset attributes | | | | |
|---|---|---|---|---|
| Data collection method with frame properties | Total number of extracted frames | Processing Method | Environment | Time and weather conditions |
| M-1 Camera mounted at a fixed place<br><br>96 dpi (horizontal and vertical resolution) with 640x480 image dimension | 8,140 | Shutterless, AGC, BPR, TD | Roadside | Daytime with cloudy weather |
| | 680 | | Alleyway | Evening time cloudy weather |
| | 4,790 | | Roadside | Night-time with light cloudy and windy weather |
| M-2 Camera mounted on the car (Driving condition)<br><br>96 dpi (horizontal and vertical resolution) with 640x480 image dimension | 9,600 | Shutterless, AGC, BPR, TD | Industrial Park | Daytime with clear weather and light foggy weather |
| | 11,960 | | Downtown | Evening time with partially cloudy and windy weather |
| | 4,600 | Shutterless, AGC, BPR, & TD | Downtown | Night-time with clear weather conditions |
| *frames* | *Daytime: 17,740 (44.61%)* | *Evening time: 12,640 (31.78%)* | *Night-time: 9,390 (23.61%)* | *Total: 39,770* |

Fig. 11 shows the six distinct sample of thermal frames captured in different environmental and weather conditions using M1 and M2 methods. These samples show different class objects such as buses, bicycles, poles, person, and cars. Most of these objects are found commonly on the roadside thus providing the driver a comprehensive video analysis of car surroundings.

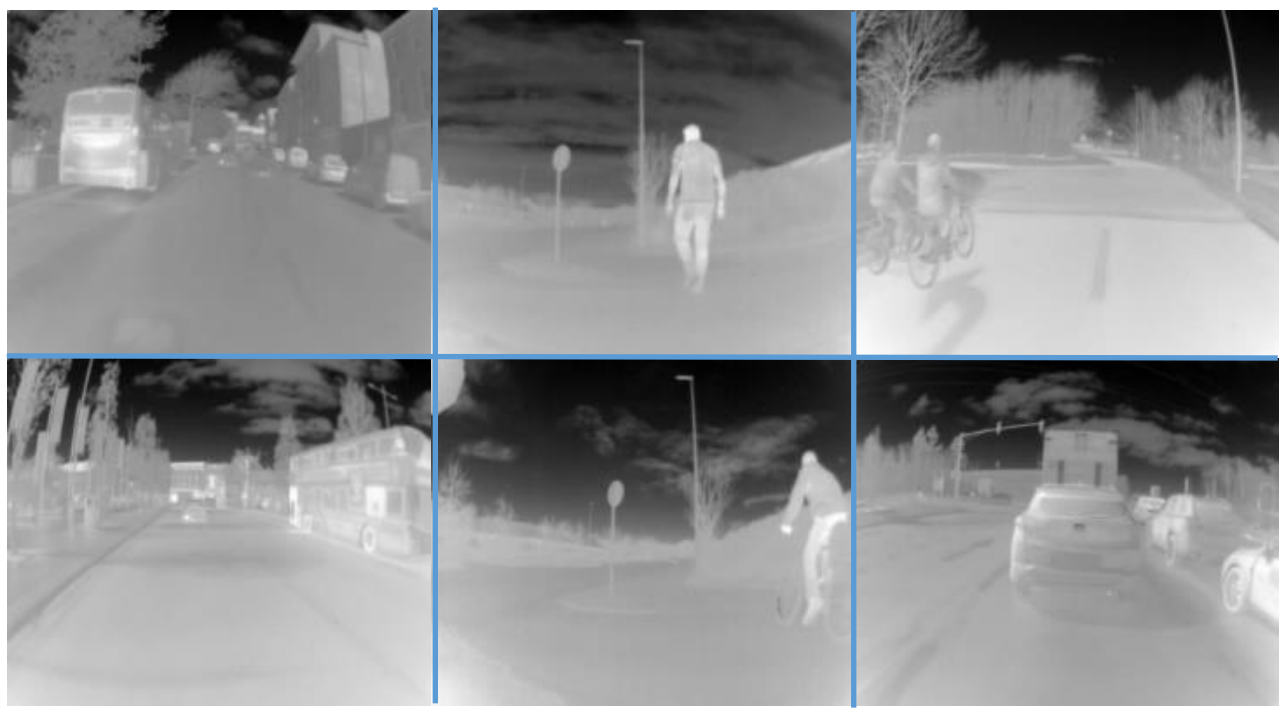

**Fig. 11.** Six different thermal samples acquired using LWIR 640x480 prototype thermal camera showing various class objects.

## IV. Proposed Methodology

This section will detail the proposed methodology and training outcomes from the various network variants tested in this study.

### *A. Network Training and Learning Perspectives*

The overall training data comprises both locally and publicly available datasets. The complete training data is divided in the ratio of 50% - 50% where 50% of data is selected from locally acquired thermal frames whereas the rest 50% of the training data leverages from public datasets. Six distinct types of roadside objects for driving assistance are included in training and validations sets. These include bicycles, motorcycles, buses, cars, pedestrians or people, and static roadside objects such as poles or road signs, as shown in Fig 12.

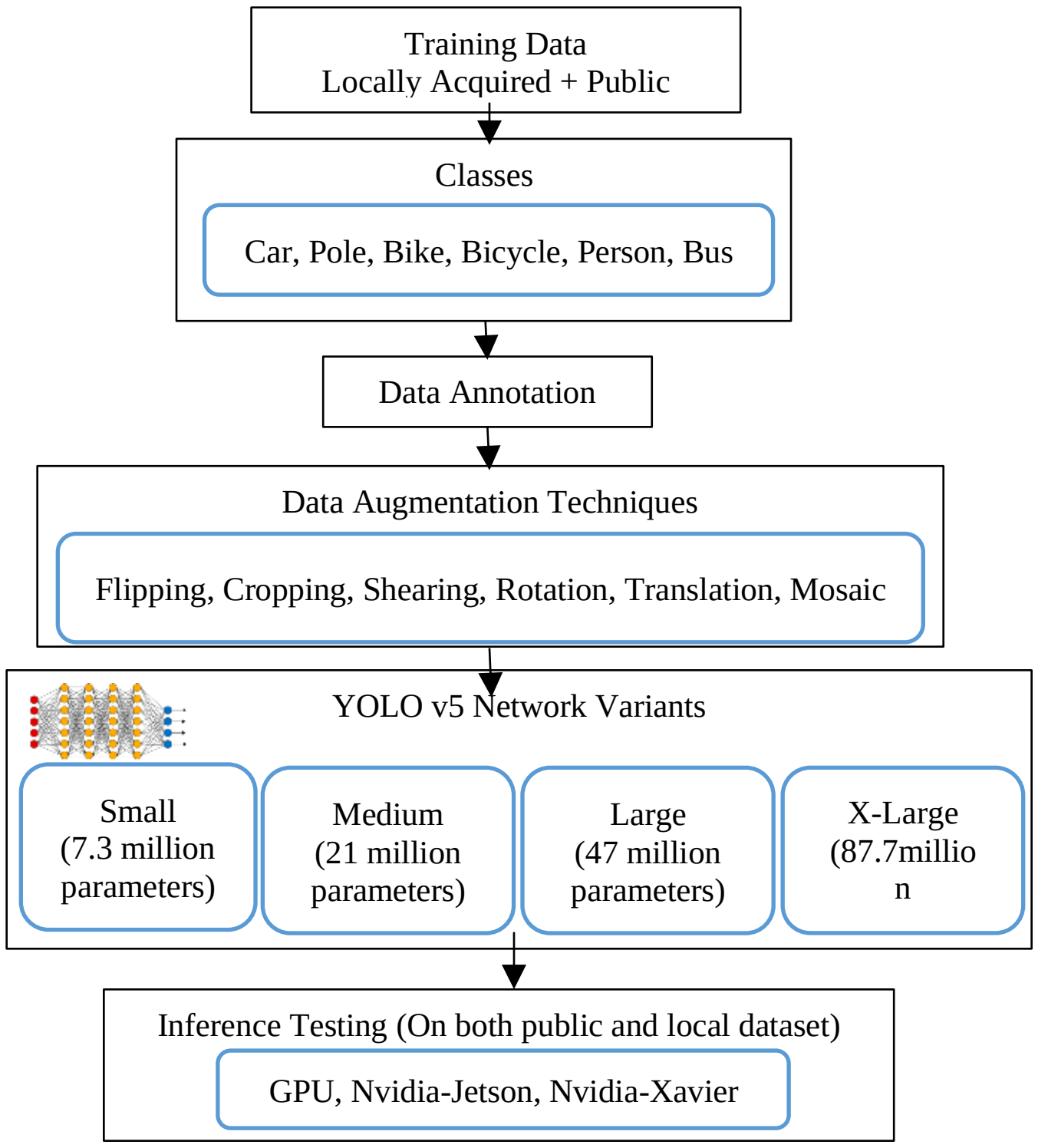

**Fig. 12.** Block diagram depicts the steps taken to evaluate the performance of Yolo v5 on local and public datasets.

Fig. 13 shows the class-wise data distribution. In the training phase of the YOLO-V5 framework, a total of 59,150 class-wise data samples were utilized, along with their corresponding class labels

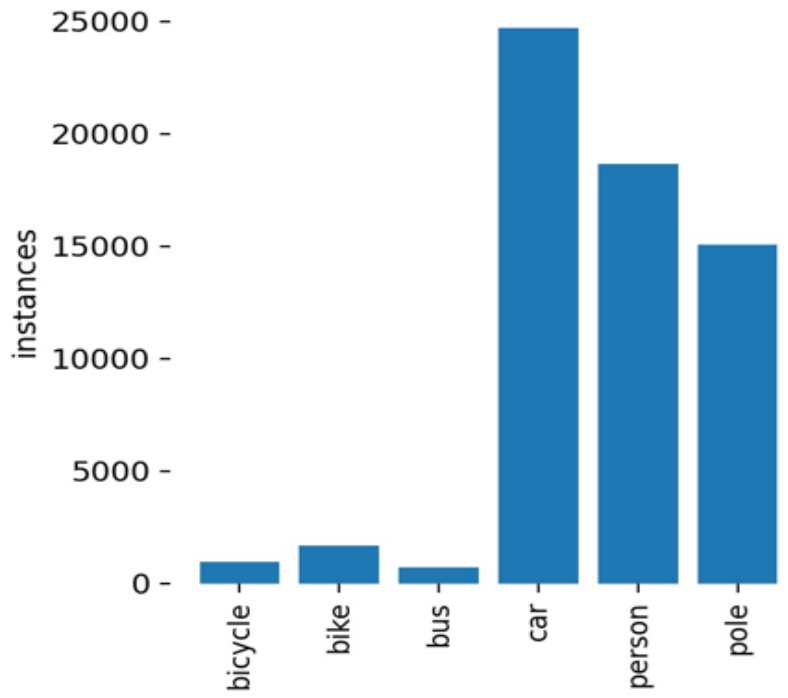

**Fig. 13.** Depicts the respective class-wise training samples distributions.

### *B. Data Annotation and Augmentation*

The overall data annotations were performed manually using an open-source bounding box-based annotations tool LabelImg [31] for all the thermal classes in our study. Annotations are stored in YOLO format as text files. During the training phase all the YoloV5 network variations which include small, medium, large, and x-large networks have been trained to detect and classify six different classes in different environmental conditions.

Large-scale datasets are considered a vital requirement for achieving optimal training results using deep learning architectures. Without the need of gathering new data, data augmentation allows us to significantly improve the diversity of data available that can be effectively used for training the DNN models. In the proposed study we have incorporated a variety of data augmentation techniques which involve cropping, flipping, rotation, shearing, translation, mosaic transformation for an optimum training of all the network variants of the YOLO-V5 framework.

### *A. C. Training Results*

As discussed in subsection A of section IV all the networks are trained using the combination of public as well as the locally gathered dataset. Training data from public datasets are included from four different datasets which include FLIR [7], OST [2], CVC [19], and KAIST [5] datasets. Secondly, we have used thermal frames acquired from the locally gathered video sets using both M1 and M2 methods. The training process is performed on a server-grade machine with XEON E5-1650 v4 3.60 GHz processor, 64 GB of ram, and equipped with GEFORCE RTX 2080 Ti graphical processing unit. It comes with 12 GB of dedicated graphical memory, memory bandwidth of 616 GB/second, and 4352 cuda cores. During the training phase the batch size is fixed to 32 and as an optimizer, both stochastic gradient descent (SGD) and ADAM optimizer were used. However, we were unable to achieve satisfactory training results using ADAM optimizer as compared to SGD thus selected SGD optimizer for training purposes. Table III shows the performance evaluation of all the trained models in the form of mean average precision (mAP), recall rate, precision, and losses.

TABLE III
TRAINING RESULTS

| Optimizer: SGD (best model *) | | | | | | |
|---|---|---|---|---|---|---|
| Network | P % | R% | mAP % | Box Loss | Object Loss | Classification Loss |
| Small | 75.58 | 65.75 | 70.71 | 0.032 | 0.034 | 0.0017 |
| Medium | 71.06 | 64.74 | 65.34 | 0.027 | 0.030 | 0.0013 |
| Large * | 82.29 | 68.67 | 71.8 | 0.025 | 0.0287 | 0.0011 |
| X-Large | 74.23 | 65.03 | 64.94 | 0.025 | 0.0270 | 0.0010 |

By analysing Table III, it can be observed that the large model performed significantly better when compared to other models with an overall precision of 82.29%, recall rate of 68.67%, and mean average precision of 71.8% mAP. Fig. 14 shows the graph results of yolo-v5 large model. The figure visualizes obtained PR-curve, box loss, object loss, and classification loss. During the training process, the X-large model consumes the maximum amount of hardware resources with the largest training time as compared to other network variants with overall GPU usage of 9.78 GB and a total training time of 14 hours. Fig. 15 shows the overall GPU memory usage, GPU power required in percentages, and GPU temperature in centigrade scale while training the largest x-large network variant of yolo-v5 model.

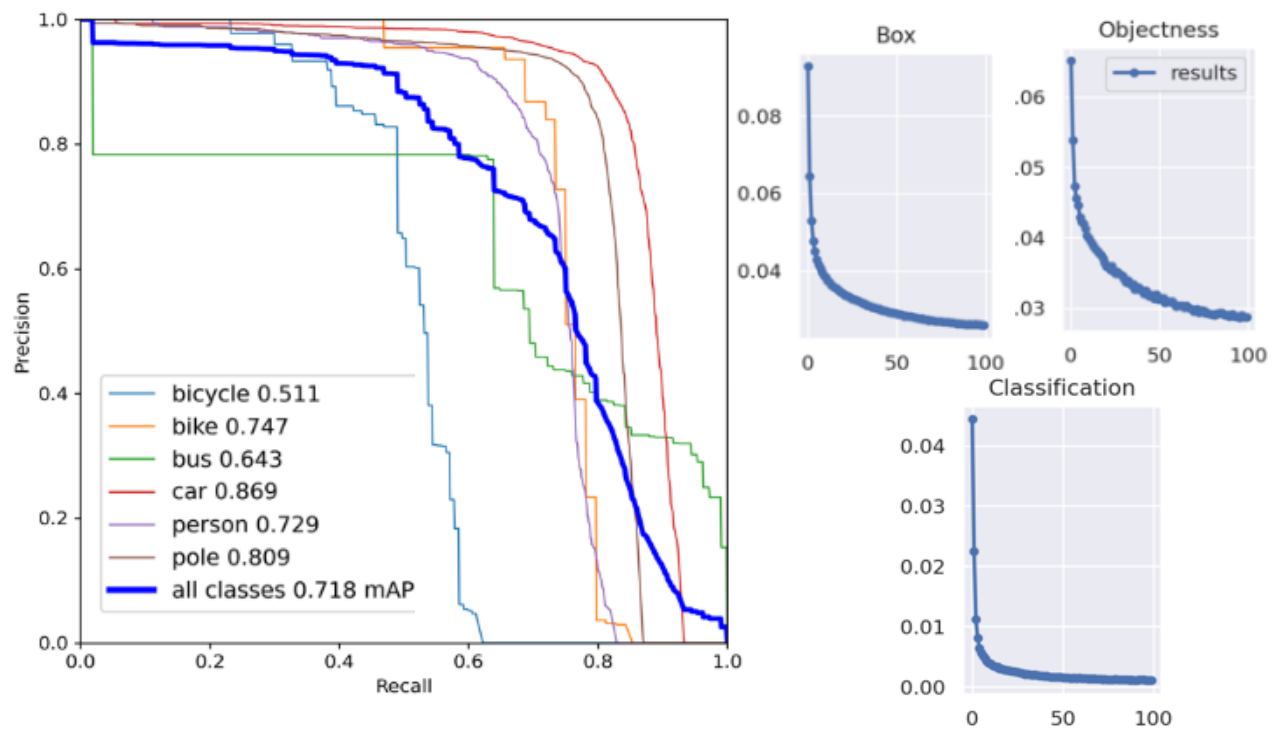

**Fig. 14.** Training results of YOLO-v5 large model using SGD optimizer.

## V. VALIDATION RESULTS ON GPU AND EDGE DEVICES

This section will demonstrate the object detection validation results on GPU as well as on two different embedded boards.

### *A. Testing Methodology and Overall Test Data*

In this research study, we have used three different testing approaches which include the conventional test-time method with no augmentation (NA), test-time augmentation (TTA), and test-time with model ensembling (ME). TTA is an extensive application of data augmentation applied to the test dataset. It performs by creating multiple augmented copies of each image in the test set, having the model make a prediction for each, then returning an ensemble of those predictions. However, since the test dataset is enlarged with a new set of augmented images the overall inference time also increases as compared to NA which is one of the downsides of this approach. TTME or ensemble learning refers to as using multiple trained networks at the same time in a parallel manner to produce one optimal predictive inference model [35]. In this study, we have tested the performance of individually trained variants of the Yolo-V5 framework and selected the best combination of models which in turn helps in achieving better validation results.

After training all the networks variants of yolo-v5, the performance of each model is cross-validated on a comprehensive set of test data selected from the public as well as locally gathered thermal data. Table IV provides the numeric data distribution of the overall validation set.

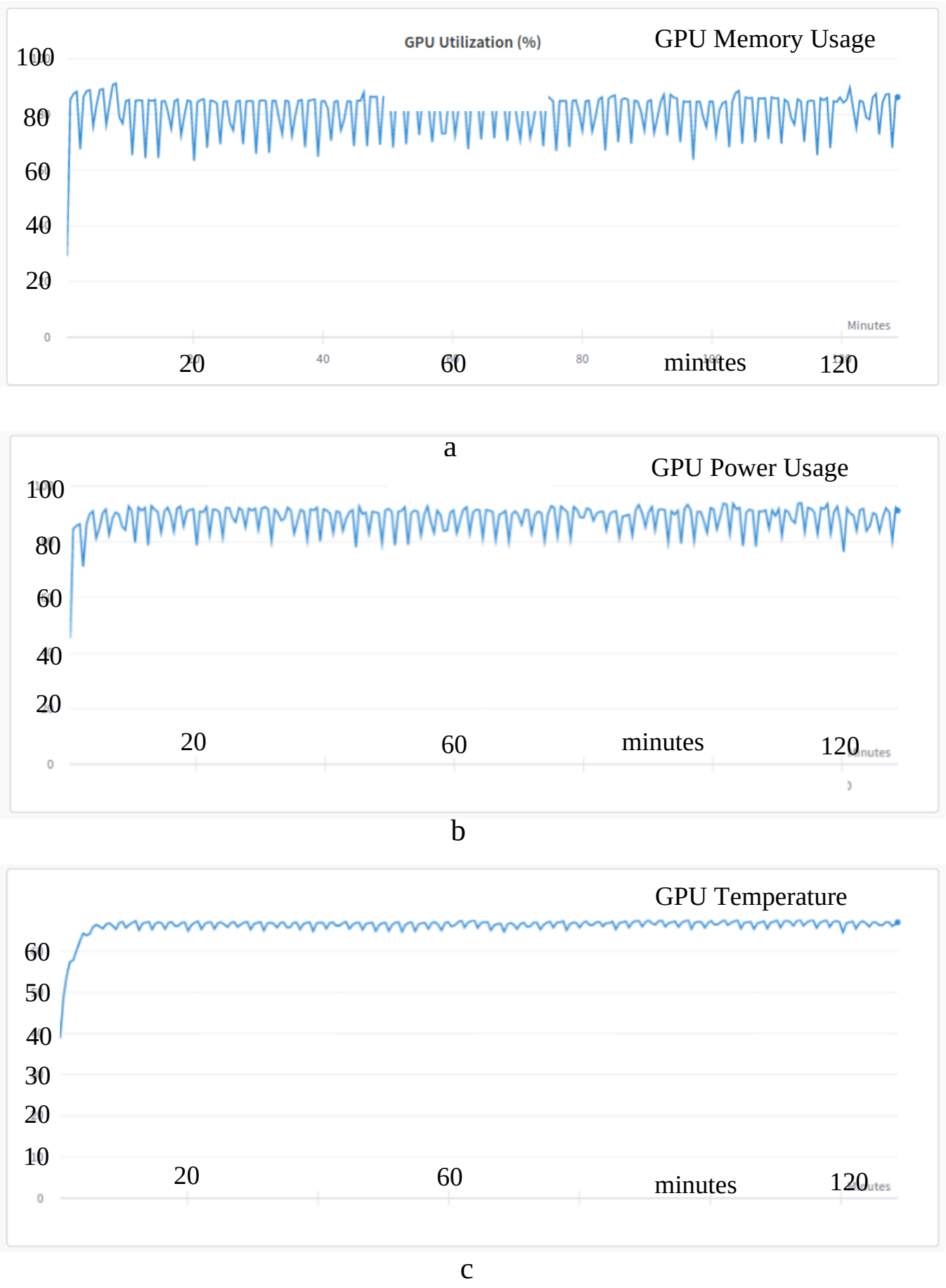

**Fig. 15.** GPU resource utilization during the training process of x-large network, a) 85% (9.78 GB) of GPU memory utilized, (b) 90% (585 watts) of GPU power required and, (c) 68 C of GPU temperature with the maximum rating of 89 C.

TABLE IV
TEST DATASET

| Test Dataset Attributes | | | | | |
|---|---|---|---|---|---|
| Frames Used | | | | | |
| Public dataset | OST | CVC-09 | KAIST | FLIR | Total No frames |
| | 50 | 5360 (day + night-time) | 149 | 130 | 5,689 |
| Local dataset | Method (M1) | | Method (M2) | | Total No frames |

| | | | |
|---|---|---|---|
| | | | |
| | 8,820 | 16,560 | 25,380 |
| | | | Total: 31,069 |

### B. *Inference Results Using YOLO Network Variants*

In the first phase, we have run the rigorous inference test on GPU as well as Edge-GPU platforms on our test data using the newly trained networks variants of yolo framework. The overall test data is consisting of nearly ≈ 31,000 thermal frames. Fig. 16 shows the inference results on 9 different thermal frames selected from both public as well as locally acquired data. These frames have data complications such as multiple class objects, occlusion, overlapping classes, scale variation, and varying environmental conditions. The complete inference results are available on our local repository (https://bit.ly/3lfvxhd).

In the second phase, we have run the combination of different models in a parallel manner using the model ensembling approach to output one optimal predictive engine which can be further used to run the inference test on the validation set. The different combination of these models is shown in Table V respectively where 1 indicates that model is in active state and 0 means model is in a non-active state.

TABLE V
MODEL ENSEMBLING

| Model Combinations | | | | | |
|---|---|---|---|---|---|
| No | Small | Medium | Large | X-Large | Combination |
| State 1 (active) or 0 (not active) | | | | | |
| 1 | *1* | *1* | *0* | *0* | A0 |
| 2 | *1* | *0* | *1* | *0* | A1 |
| 3 | *1* | *0* | *0* | *1* | A2 |
| 4 | *0* | *1* | *1* | *0* | A3 |
| 5 | *0* | *0* | *1* | *1* | A4 |

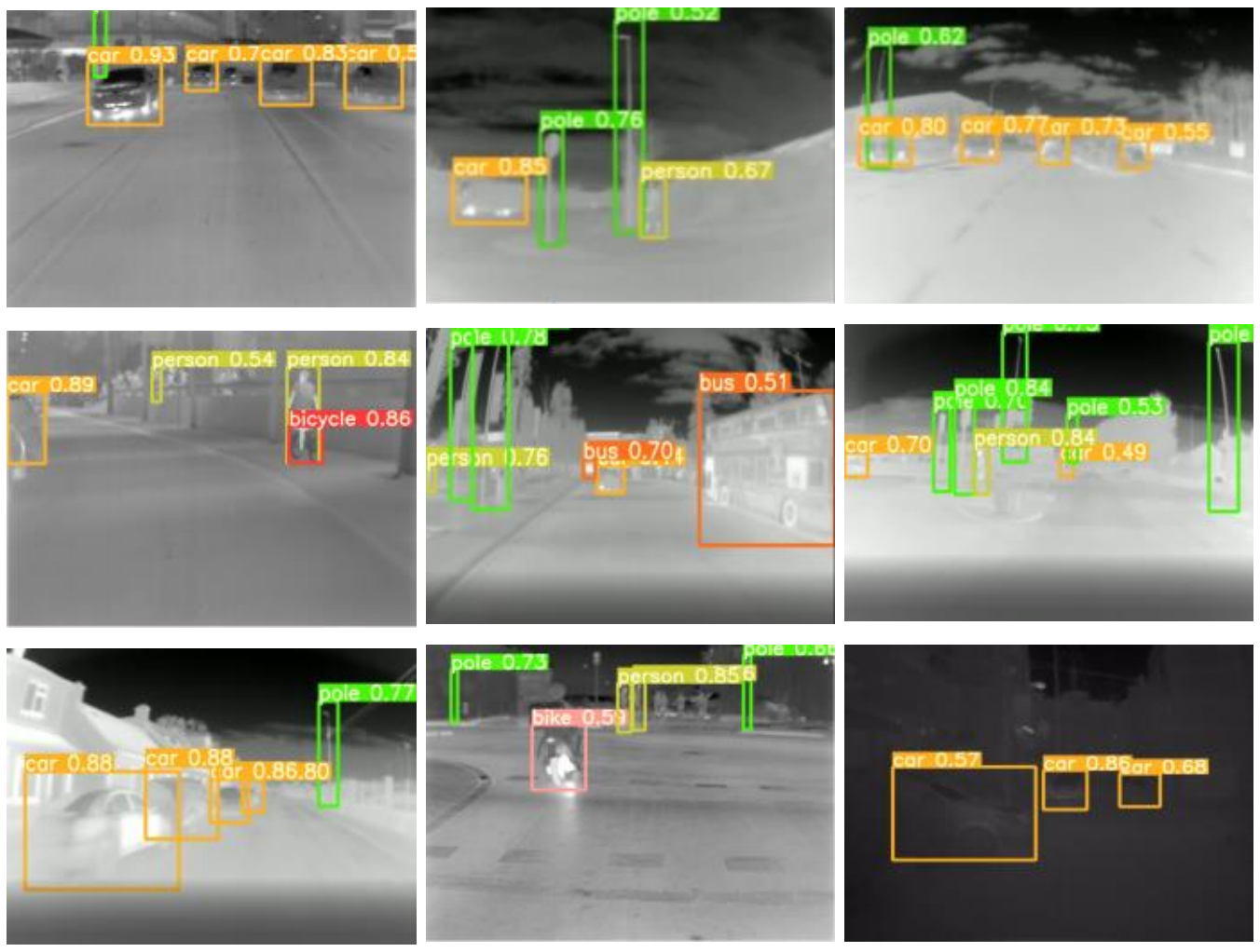

**Fig. 16**. Inference results on nine different frames selected from test data.

With the model ensembling method small and large models (A1) turn out to best model combination in terms of achieving the best mAP, recall, and relatively less amount of inference time per frame thus producing optimal validation results. These results are examined in further parts of this section. Fig. 17 shows the inference results using A1 model ensembling engine on three different thermal frames selected from the test data.

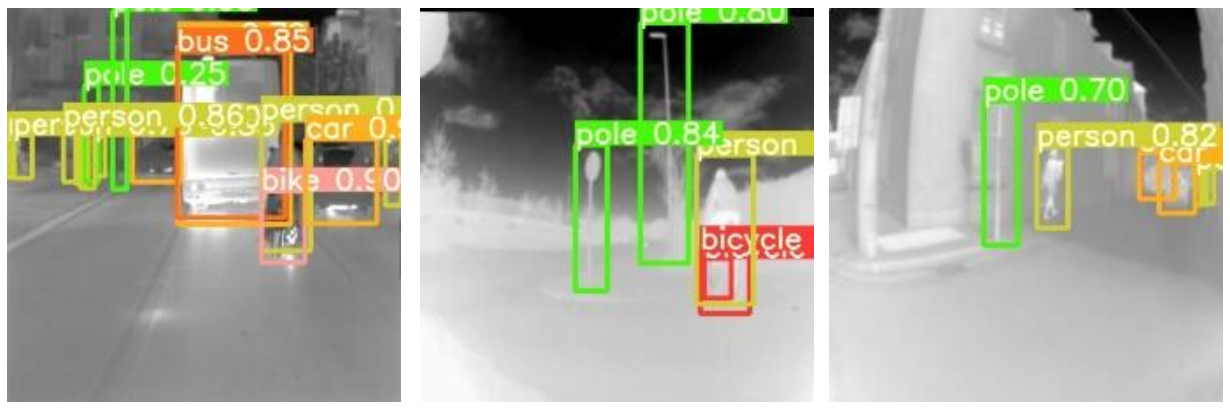

**Fig. 17.** Inference results on three different frames using model ensembling.

### C. *Quantitative Validation Results on GPU*

The third part of the testing phase shows the quantitative numerical results of all the trained models on GPU. To better analyse and validate the overall performance for all the trained models on test data, relatively a smaller set of test images has been selected from the overall test set. For this purpose, a subset of 402 thermal frames is selected to compute all the evaluation metrics. The selected images consist of different roadside objects such as pedestrians, cars and buses under different illumination and environmental conditions, time of day, and distance from the camera. The objects are either far-field (between 11-18 meters), mid-field (between 7-10 meters) or near-field (between 3-6 meters) from the camera. Fig. 18 shows selected views from the test data for quick reference of the reader.

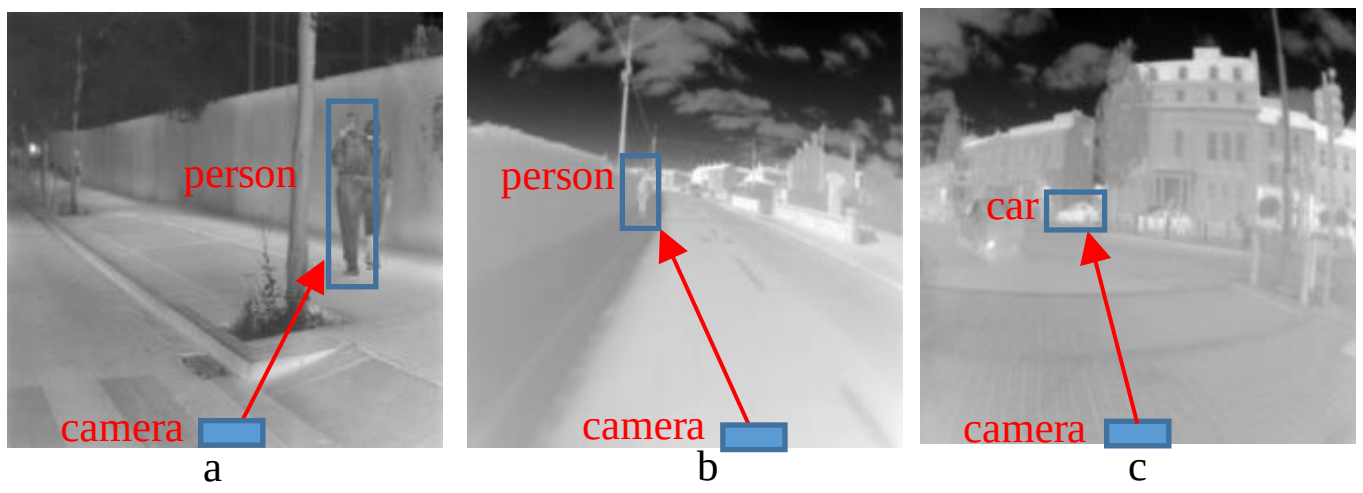

**Fig. 18.** Test data samples with the object at varying distances from the camera, (a) near-field distance, (b) mid-field distance, (c) far-field distance.

The performance evaluation of each model is computed using four different metrics which include recall, precision, mean average precision (mAP), and frames per second rate (FPS). Table VI shows all the quantitative validation results on GPU. During the testing phase batch size is fixed to 8. Also, three different testing configuration is selected thus having separate confidence threshold values and the intersection of union values at each validation phase. Confidence threshold defines the minimum threshold value, or in other words, it is the minimum confidence score above which we consider a prediction as true. If it's below the threshold value, we consider the prediction as "no". The last row of Table VI shows the best ME results using A1 configuration from Table V with a selected confidence threshold of 0.2 and IoU threshold of 0.4.

TABLE VI
QUANTITATIVE RESULTS ON GPU

| Platform: GPU<br>Inference image size: 800 x 800 | | | | | | | | |
|---|---|---|---|---|---|---|---|---|
| Confidence Threshold: 0.4, IoU Threshold: 0.6 | | | | | | | | |
| No Augmentation (NA) | | | | | Test-time Augmentation (TTA) | | | |
| Network | P % | R % | mAP% | FPS | P % | R % | mAP% | FPS |
| Small | 72 | 46 | 43 | 79 | 76 | 48 | 50 | 45 |
| Medium | 73 | 54 | 49 | 53 | 76 | 58 | 57 | 26 |
| Large | 75 | 56 | 52 | 34 | 77 | 63 | 60 | 16 |
| X-Large | 74 | 53 | 49 | 20 | 71 | 59 | 55 | 10 |
| Confidence Threshold: 0.2, IoU Threshold: 0.4 | | | | | | | | |
| NA | | | | | TTA | | | |
| Small | 66 | 50 | 47 | 82 | 64 | 55 | 52 | 45 |
| Medium | 66 | 57 | 51 | 53 | 77 | 58 | 59 | 27 |
| Large | 71 | 61 | 56 | 35 | 78 | 63 | 63 | 16 |
| X-Large | 70 | 54 | 50 | 21 | 68 | 62 | 56 | 10 |
| Confidence Threshold: 0.1, IoU Threshold: 0.2 | | | | | | | | |
| NA | | | | | TTA | | | |
| Small | 65 | 52 | 48 | 81 | 65 | 53 | 53 | 45 |
| Medium | 69 | 54 | 51 | 53 | 77 | 58 | 59 | 26 |
| Large | 73 | 61 | 57 | 34 | 79 | 63 | 63 | 16 |
| X-Large | 71 | 54 | 52 | 21 | 69 | 62 | 57 | 10 |
| Confidence Threshold: 0.2, IoU Threshold: 0.4 | | | | | | | | |
| Model Ensembling (ME) | | | | | | | | |
| A = Small<br>B = Large<br>Comb: A1<br>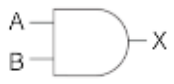 | --- | --- | --- | --- | 77 | 66 | 65 | 25 |

### D. *Quantitative Validation Results on Edge-GPU Devices*

This section will review the quantitative validation results on two different Edge-GPU platforms (Jetson Nano & Jetson Xavier NX). It is pertinent to mention that Jetson Xavier NX development kit embeds more computational power in terms of GPU, CPU, and memory as compared to Nvidia Jetson Nano. Table VII shows the specification comparison of both boards.

TABLE VII
HARDWARE SPECIFICATION COMPARISON

| Hardware specification comparison of Nvidia Jetson Nano and Nvidia Jetson Xavier NX | | |
|---|---|---|
| Board | Jetson Nano [23] | Jetson Xavier NX [25] |
| CPU | Quad-Core ARM® Cortex® -A57 MPCore, 2 MB L2, Maximum Operating Frequency: 1.43 GHz | 6-core NVIDIA Carmel ARM®v8.2 64-bit CPU, 6 MB L2 + 4 MB L3, Maximum Operating Frequency: 1.9 GHz |
| GPU | 128-core Maxwell GPU, 512 GFLOPS (FP16), Maximum Operating Frequency: 921 MHz | 384 CUDA® cores + 48 Tensor cores Volta GPU, 21 TOPS, Maximum Operating Frequency: 1100 MHz |
| RAM | 4 GB 64-bit LPDDR4 @ 1600MHz \| 25.6 GB/s | 8 GB 128-bit LPDDR4x @ 1600MHz \| 51.2GB/s |
| On module Storage | 16 GB eMMC 5.1 Flash Storage, Bus Width: 8-bit, Maximum Bus Frequency: 200 MHz (HS400) | |
| Thermal Design Power | 5W – 10W | 10W – 15W |
| AI Performance | 0.5 TFLOPS (FP16) | 6 TFLOPS (FP16)<br>21 TOPS (INT8) |

On Jetson Nano we have validated the performance of the small version only whereas on Jetson Xavier NX we have evaluated the performance of smaller and medium versions of models due to the memory limitations and constrained hardware resources on these boards. During the testing phase, we have selected the highest power modes on both boards to provide the utmost efficiency thus utilizing maximum hardware resources. For instance, on Nvidia Xavier board NX we have selected 'Mode Id: 2' which means the board is operating in 15-watt power mode with all the six cores active with a maximal CPU frequency of 1.4 gigahertz and GPU frequency of 1.1 gigahertz. Similarly, on Nvidia Jetson Nano all the four CPU cores were utilized with overall power utilization of 5 watts. Table VIII shows the quantitative validation results on ARM processor based embedded boards

TABLE VIII
QUANTITATIVE RESULTS ON EDGE PLATFORMS

| Platform: Nvidia Jetson Nano<br>Inference image size: 128 x 128 | | | | | | | | |
|---|---|---|---|---|---|---|---|---|
| Confidence Threshold: 0.4, IoU Threshold: 0.6 | | | | | | | | |
| | NA | | | | TTA | | | |
| | P % | R % | mA P% | FPS | P % | R % | mA P% | FPS |
| Small | 75 | 44 | 45 | 3 | 77 | 47 | 49 | 1 |
| Confidence Threshold: 0.2, IoU Threshold: 0.4 | | | | | | | | |
| | NA | | | | TTA | | | |
| Small | 75 | 44 | 47 | 3 | 71 | 51 | 51 | 1 |
| Confidence Threshold: 0.1, IoU Threshold: 0.2 | | | | | | | | |
| | NA | | | | TTA | | | |
| Small | 66 | 47 | 48 | 2 | 73 | 50 | 52 | 1 |
| Platform: Nvidia Jetson Xavier NX<br>Inference image size: 128 x 128 | | | | | | | | |
| Confidence Threshold: 0.4, IoU Threshold: 0.6 | | | | | | | | |
| | NA | | | | TTA | | | |
| Small | 75 | 44 | 45 | 18 | 77 | 47 | 49 | 10 |
| Med | 76 | 53 | 50 | 12 | 79 | 50 | 52 | 6 |
| Confidence Threshold: 0.2, IoU Threshold: 0.4 | | | | | | | | |
| | NA | | | | TTA | | | |
| Small | 75 | 44 | 47 | 19 | 71 | 51 | 51 | 10 |
| Med | 76 | 52 | 53 | 12 | 73 | 54 | 53 | 6 |
| Confidence Threshold: 0.1, IoU Threshold: 0.2 | | | | | | | | |
| | NA | | | | TTA | | | |
| Small | 66 | 47 | 48 | 18 | 73 | 50 | 52 | 10 |
| Med | 76 | 51 | 52 | 12 | 81 | 49 | 53 | 6 |

### E. *Real-time Hardware Feasibility Testing*

While running these tests we closely monitor the temperature ratings of different hardware peripherals on both Edge-GPU platforms. It is done to prevent the overheating effect which can damage the onboard processor or effect the overall operational capability of the system. In the case of Nvidia Jetson Nano, a cooling fan was mounted on top of the processor heatsink to reduce the overheating effect as shown in Fig. 19.

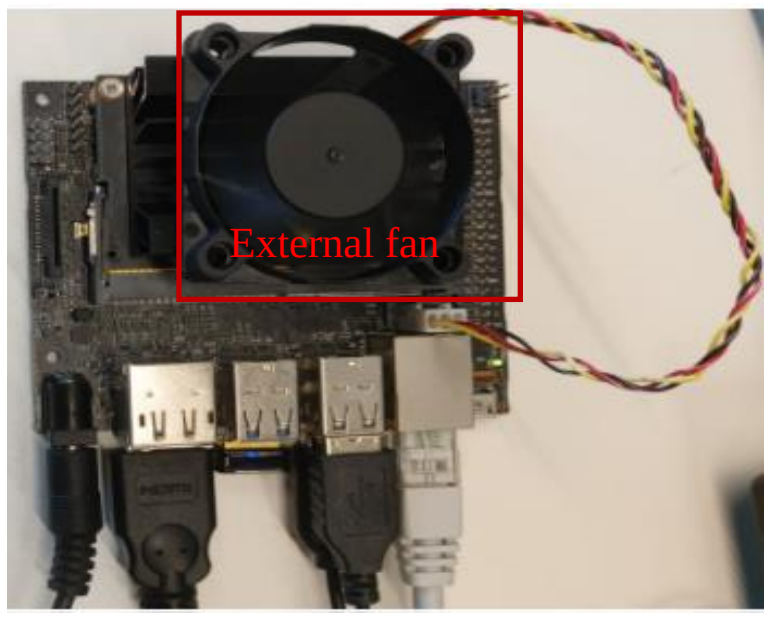

**Fig. 19.** External 5-volt fan unit mounted on Nvidia Jetson Nano processor heatsink to avoid onboard overheating effect while running the inference testing.

The temperature ratings of various hardware peripherals are monitored using eight different on-die thermal sensors and one on-die thermal diode. These temperature monitors are referred to as CPU-Thermal, GPU-Thermal, Memory-Thermal, and PLL-Thermal (part thermal zone). External fans help us in reducing the temperature rating of various hardware peripherals drastically as compared to without mounting the fan. Fig. 20 shows the temperature rating difference of onboard thermal sensors while running the smaller version of the model on Nvidia Jetson Nano without and with mounting the external cooling fan.

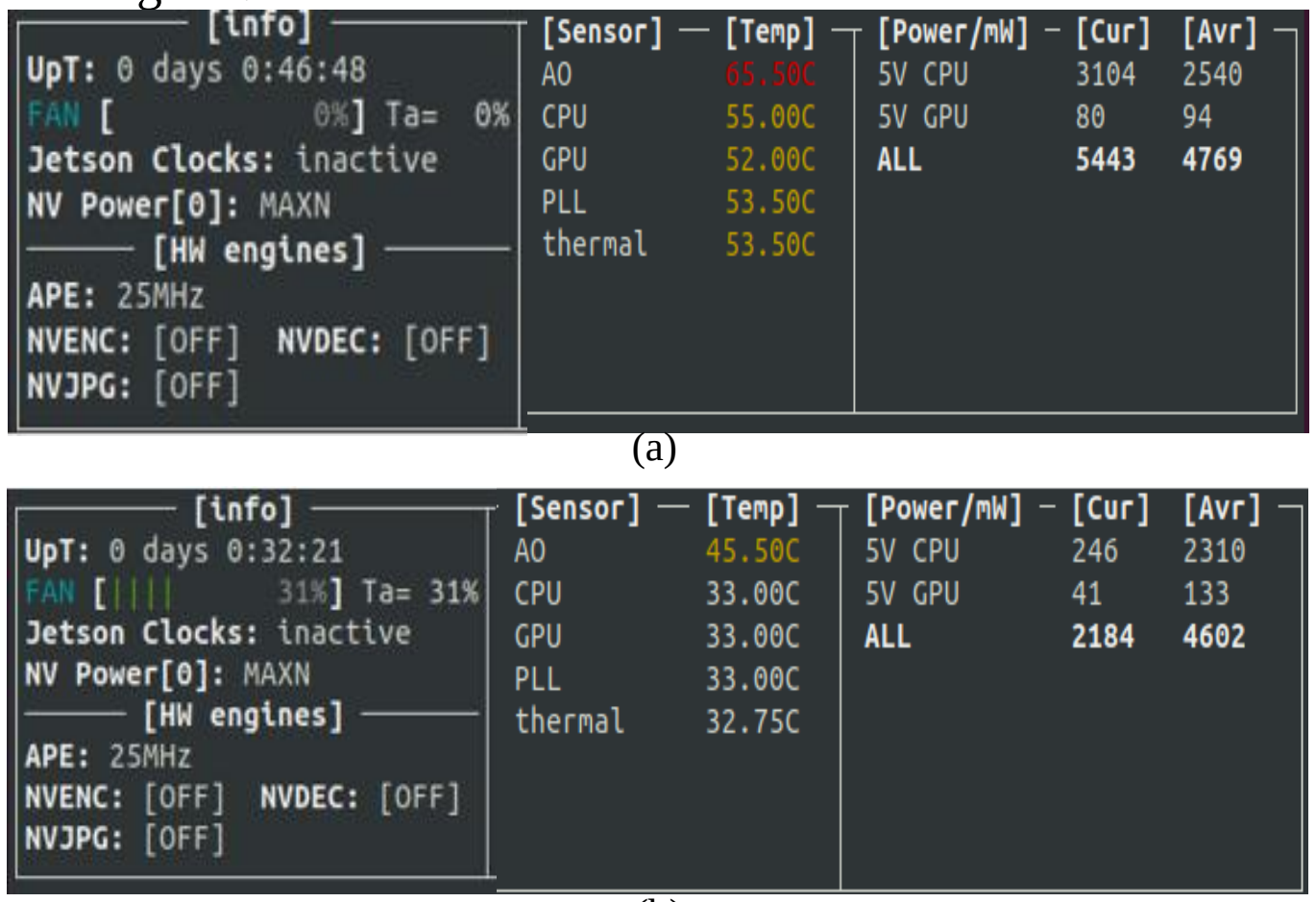

**Fig. 20.** Temperature rating difference of different onboard hardware peripherals on Jetson Nano (a) without fan: A0 thermal zone = 65.50 C, CPU = 55 C, GPU = 52 C, PLL: 53.50, overall thermal temperature = 53.50 C, (b) with external fan: A0 thermal zone = 45.50 C, CPU = 33 C, GPU = 33 C, PLL: 33, overall thermal temperature = 32.75 C.

It can be examined from Fig. 20 that by mounting an external cooling fan the temperature rating of various onboard peripheral on Jetson Nano was reduced by nearly 30% thus allowing us to operate the board at its maximum capacity for rigorous model testing. Fig. 21 shows the Nvidia Jetson running at its full pace (with an external fan) such that all the four cores running at their maximum limit (100% capacity) while running the quantitative and inference test by deploying the smaller network variant of the yolo-v5 framework.

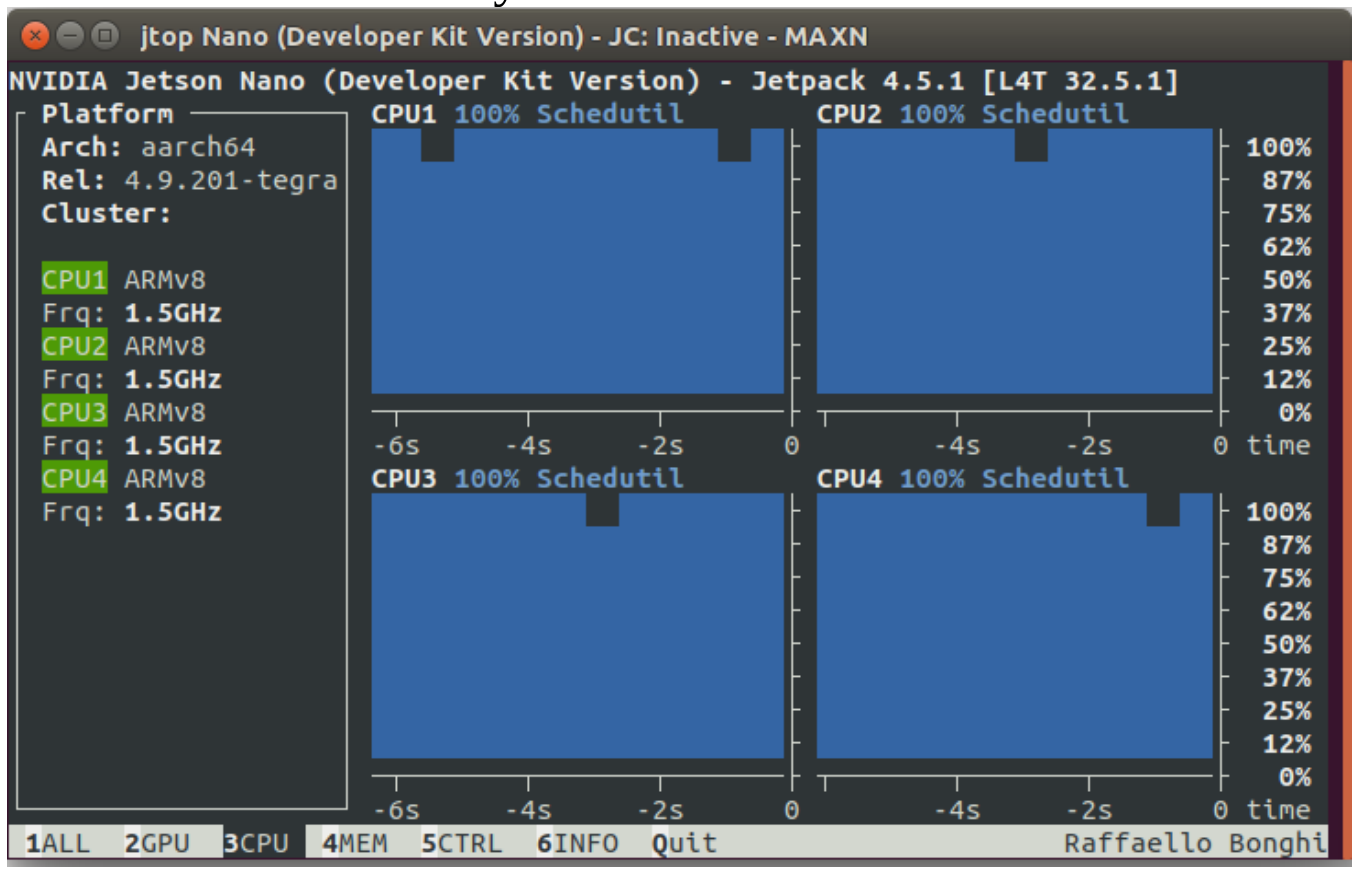

**Fig. 21.** Nvidia Jetson Nano running at MAXN power mode with all the cores running at their maximum capacity while running the inference test and quantitative validation test.

Fig. 22 shows the temperature rating difference of onboard thermal sensors while running the smaller version of the model on Nvidia Jetson Xavier NX board. Whereas Fig. 23 shows the CPU and GPU usage while running the smaller variant of YOLO-V5 framework for quantitative validation and inference test on Nvidia Xavier NX development kit.

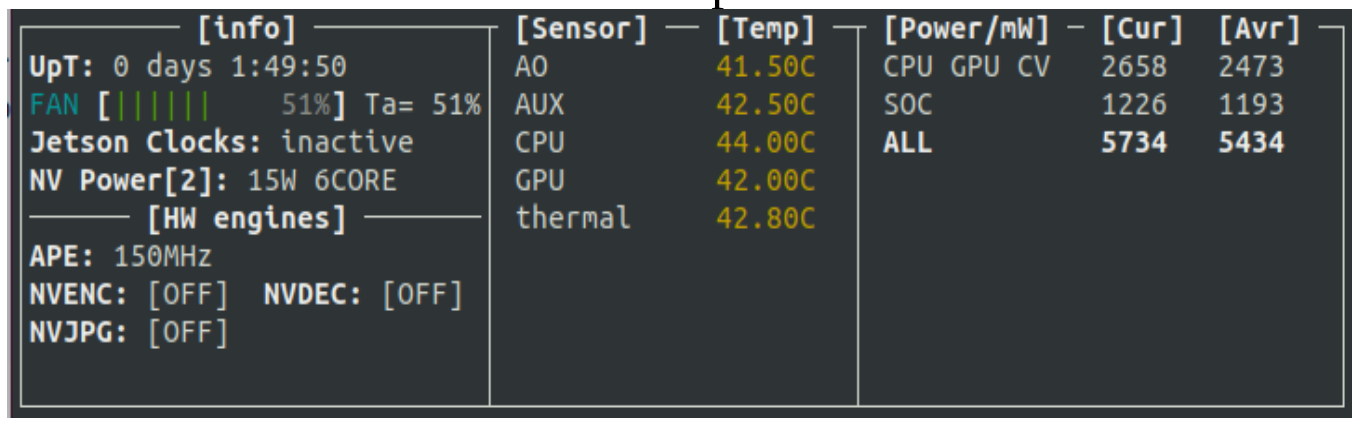

**Fig. 22.** Temperature rating of different onboard hardware peripherals on Jetson Xavier NX (a) A0 thermal zone = 41.50 C, AUX: 42.5 C, CPU = 44 C, GPU = 42 C, overall thermal temperature = 42.80 C,

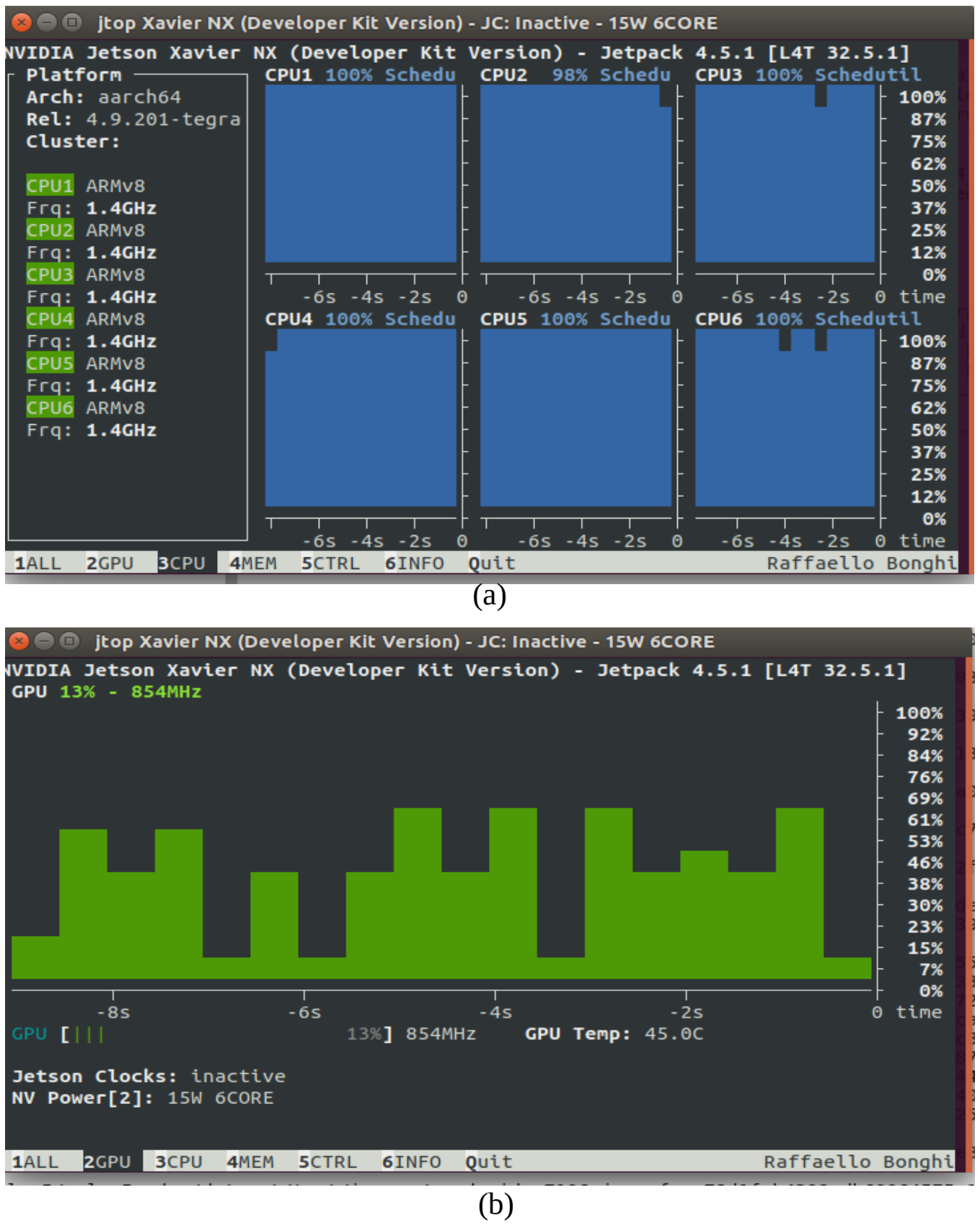

**Fig. 23.** Nvidia Jetson Xavier running at 15-watt 6 core power mode, (a) all the CPU cores running at its maximum capacity while running the quantitative validation test, (b) 69% GPU utilization while running the inference test with an image size of 128 x 128.

## VI. Model Performance Optimization(s)

This section will mainly aim at further model optimization using TensorRT [33] inference accelerator tool. The prime reason for this is to further increase the FPS rate for real-time evaluation and on-board feasibility testing on edge devices. Secondly, it helps in saving onboard memory footprints on the target device by performing various optimization methods.

TensorRT [33] works by performing five modes of optimization methods for increasing the throughput of deep neural networks. In the first step, it maximizes throughput by quantizing models to 8-bit integer data type or FP16 precision while preserving the model accuracy. This method significantly reduces the model size since it is transformed from originally FP32 to FP16 version. In the next step, it uses layer and tensor fusion techniques to further optimize the usage of onboard GPU memory. The third step includes performing kernel auto-tuning. It is the most important step where the TensorRT engine shortlists the best network layers, and optimal batch size based on the target GPU hardware. In the second last step, it minimizes memory footprints and re-uses memory by distributing memory to tensor only for the period of its usage. In the last steps, it processes multiple input streams in parallel and finally optimizes neural networks periodically with dynamically generated kernels [33].

In the proposed research work we have deployed a smaller variant of yolo-v5 using TensorRT inference accelerator on both edge platforms Nvidia Jetson Nano and Nvidia Jetson Xavier NX development boards to further excel the performance of the trained model. It produces faster inference time thus increasing the FPS on thermal data which in turn helps us in building an effective real-time forward sensing system for ADAS embedded applications. Fig. 24 depicts the block diagram representation of deployment phase TensorRT inference accelerator on embedded platforms.

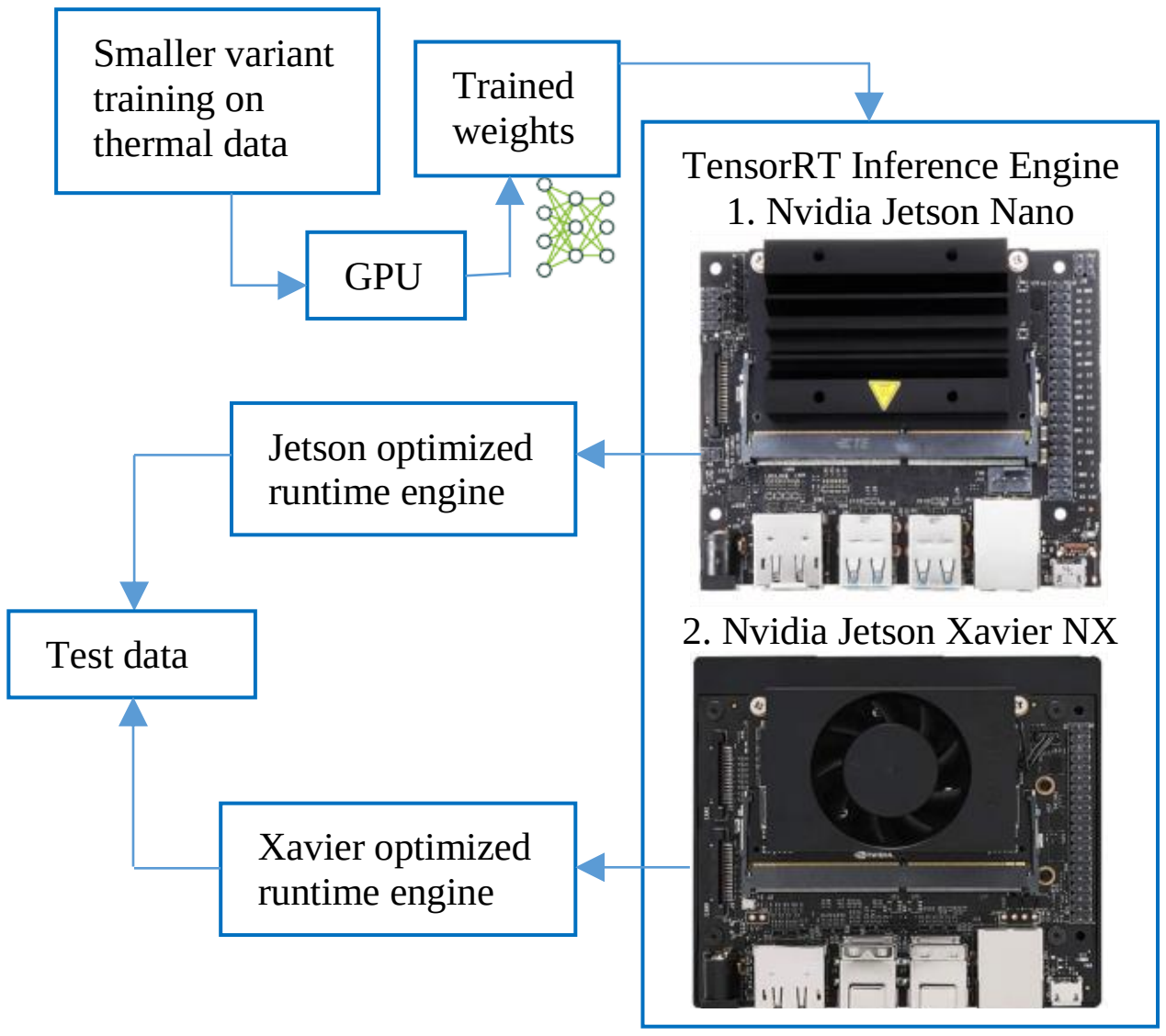

**Fig. 24.** Overall block diagram representation of deployment and running TensorRT inference accelerator on two different embedded platforms.

Table IX shows the overall inference time along with FPS rate on thermal test data using TensorRT run-time engine. By analyzing the results from Table. IX we can deduce that TensorRT API supports in boosting the overall FPS rate on ARM-based embedded platforms by nearly 3.5 times as compared to the FPS rate achieved by running the non-optimized smaller variant on Nvidia Jetson Nano and Nvidia Jetson Xavier boards. The same is demonstrated via graphical chart results in Fig. 25.

TABLE IX

TENSORRT INFERENCE ACCELERATOR RESULTS

| FPS on Nvidia Jetson Nano and Nvidia Jetson Xavier NX | | |
|---|---|---|
| Board | Nvidia Jetson Nano | Nvidia Jetson Xavier NX |
| Test Data | 402 images with the resolution of 128x128 | |
| Overall inference time | 35,090 milliseconds ≈ 35.1 seconds | 6,675 milliseconds ≈ 6.7 seconds |
| PS | 35.1 sec / 402 frames = 0.087 sec/frame<br>FPS: 1 sec / 0.087 = 11.49 ≈ 11 fps | 6.7 sec / 402 frames = 0.0166 sec/frame<br>FPS: 1 sec / 0.0166 = 60.24 ≈ 60 fps |

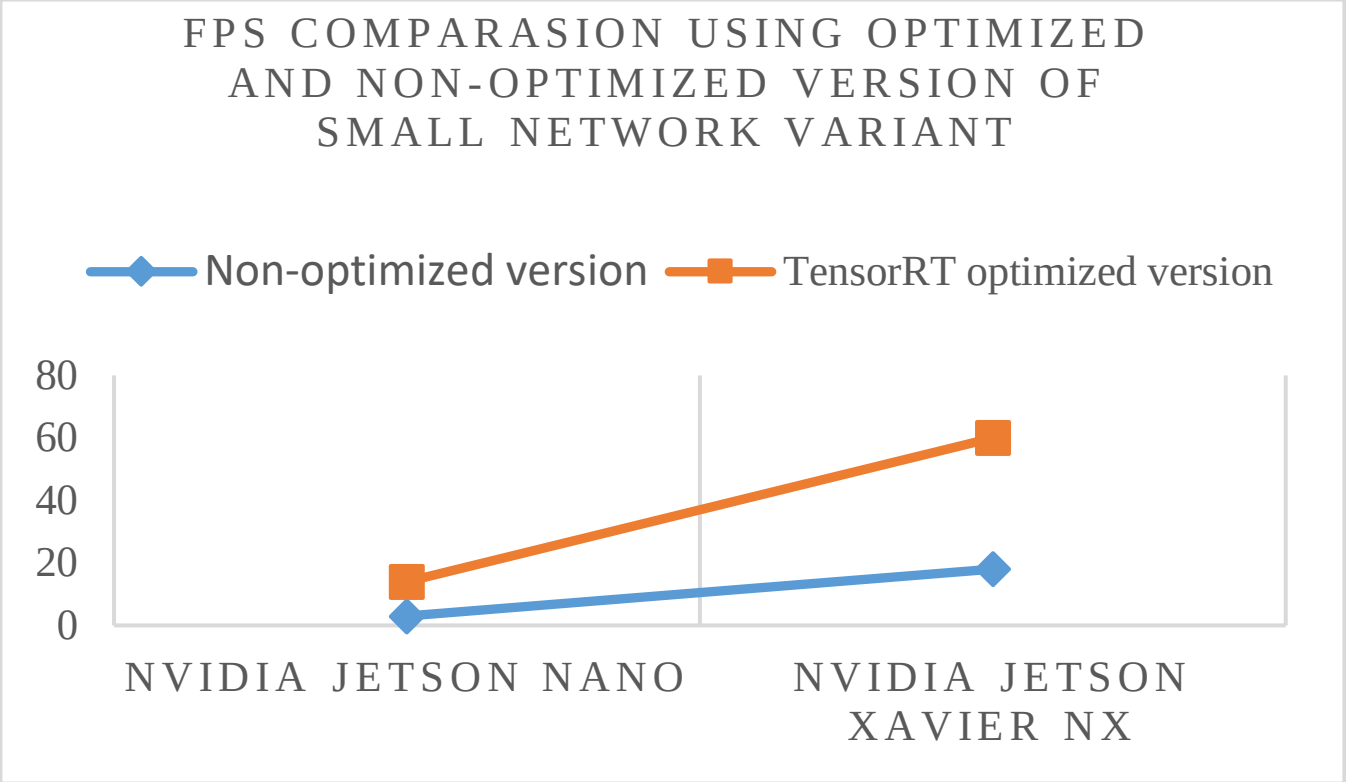

**Fig. 25.** FPS increment rate of nearly 3.5 times on Jetson Nano and Jetson Xavier NX embedded boards using the TensorRT built optimized inference engine.

Fig. 26 shows the thermal object detection inference results on six different thermal frames from the public as well as locally acquired test data produced through the neural accelerator.

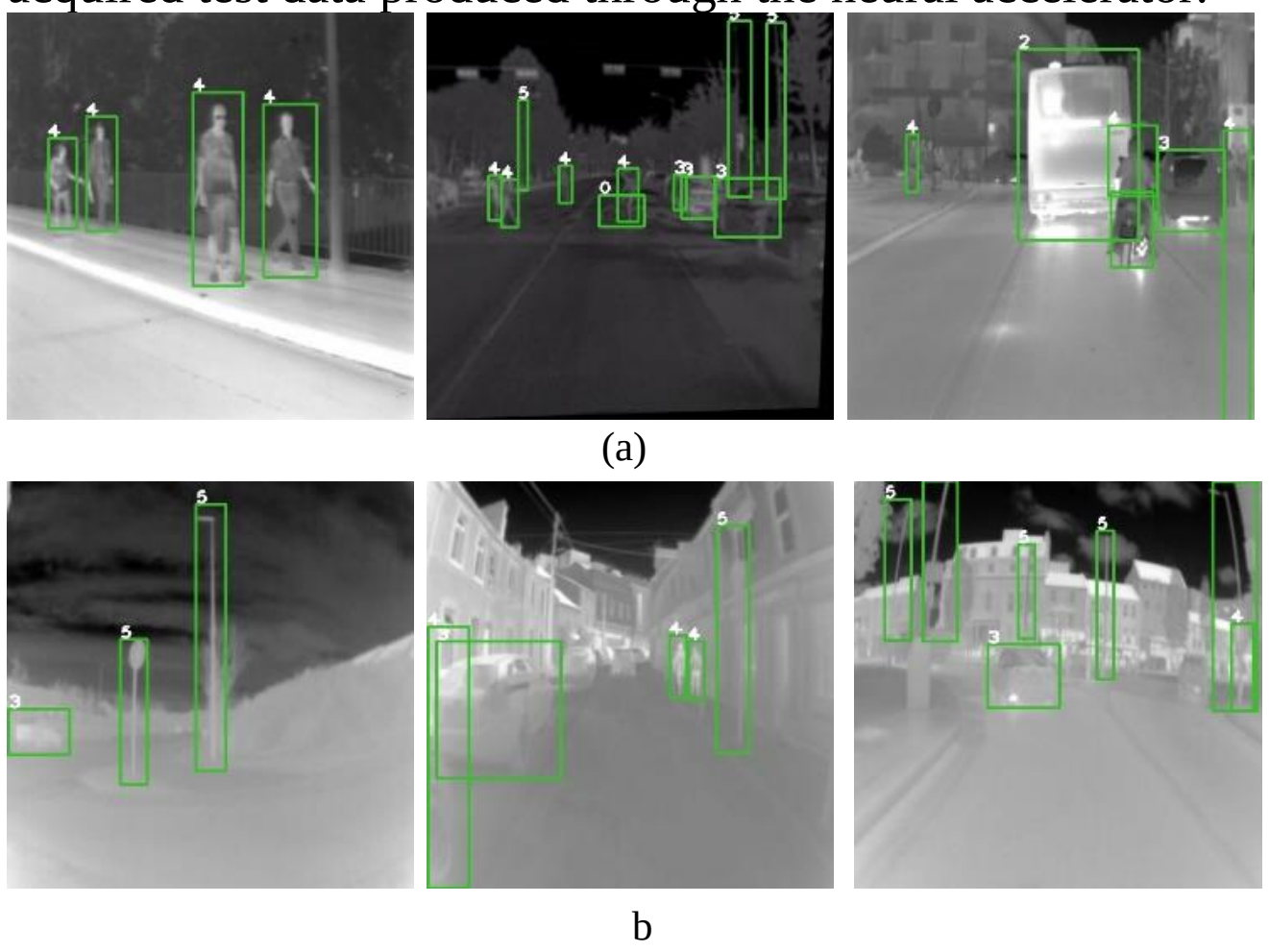

**Fig. 26.** Inference results using TensorRT neural accelerator, (a) Object detection results on public data, (b) Object Detection results on locally acquired thermal frames.

## DISCUSSION/ ANALYSIS

This section will review the training and testing performance of all YOLO-V5 framework model variants.

- During the training phase, the large YOLO v5 network outperforms other network variants scoring the highest precision of 82.29% and a mean average precision (mAP) score of 71.8%.
- Although the large network variant performed significantly better during the training phase, the small network variant

also performed well with an overall precision of 75.58% and mAP of 70.71%. Also, it gains a higher FPS rate on GPU during the testing phase as compared to the large model. Fig. 27 summarizes the quantitative performance comparison of small and large network variants of yolo framework.

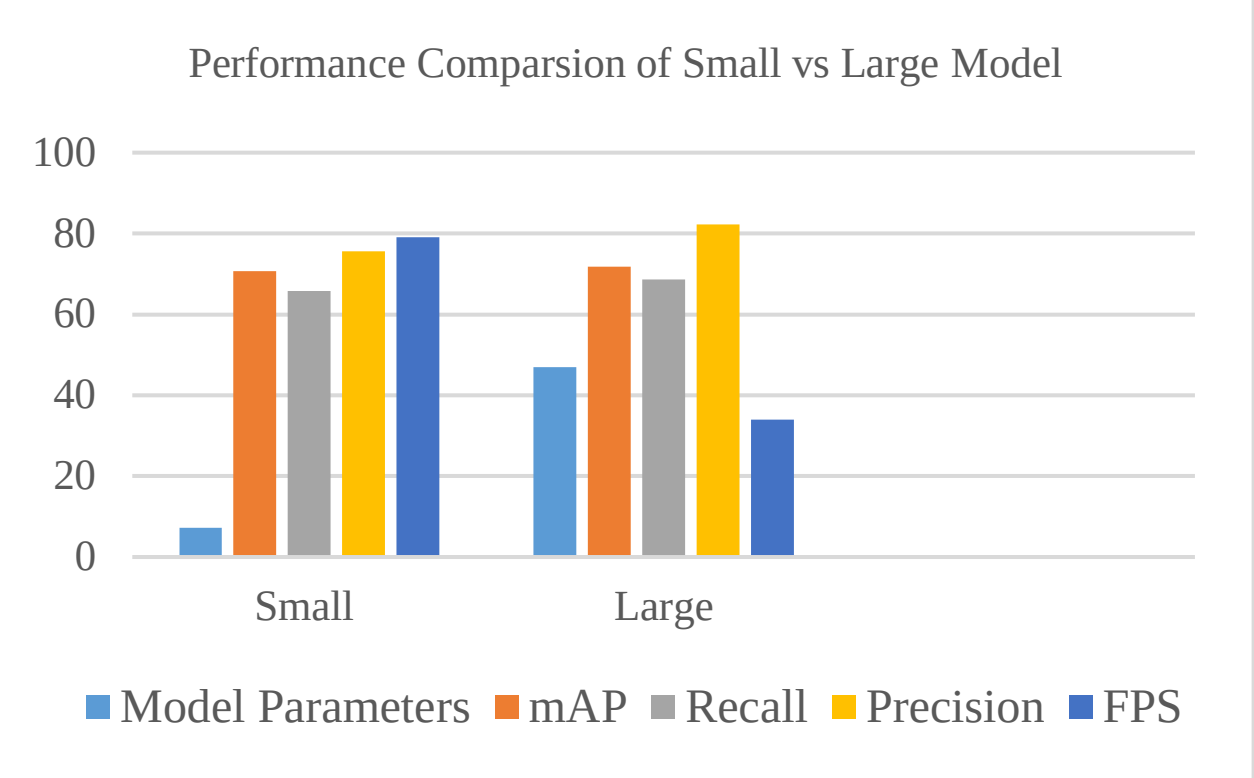

**Fig. 27.** Quantitative metrics comparison of small and large network variants

- Due to the smaller number of model parameters as compared to larger network variant (7.3M Vs 47M model parameters) and faster FPS rate on GPU during the testing phase as shown in Fig. 26 this model is shortlisted for validation and deployment purposes on both the edge embedded platforms Nvidia Jetson Nano and Nvidia Jetson Xavier NX kits.
- During the testing phase, it was noticed that by reducing the confidence threshold from 0.4 to 0.1 and the IoU threshold from 0.6 to 0.2 in three stepwise intervals, the model's mAP and recall rates increased significantly, but the precision level decreases. However, the FPS rate remains effectively constant in most of the trained model cases.
- TTA methods achieved improved testing results when compared to the NA method however the main drawback of this method is that the FPS rate drops substantially which is not suitable for real-time deployments. To overcome this problem a model ensembling (ME) based inference engine is proposed. Table IV shows the ME results by running large-small model in parallel configuration with a confidence threshold of 0.2, and an IoU Threshold of 0.4. The ensembling engine attains an overall mAP of 66% with 25 frames per second.
- When comparing the individual hardware resources of both the edge platforms (NVidia Jetson Nano and Jetson Xavier), Xavier is computationally more powerful than the Jetson Nano. Note that due to memory limitations and the lower computational power of the Jetson only the small network variant was evaluated on the Jetson Nano, whereas both the smaller and medium network variants were evaluated on the Jetson Xavier NX.
- It was observed that throughout the testing phase, it was important to keep a close eye on the operational temperature ratings of different onboard thermal sensors to avoid overheating, which might damage the onboard components or affect the system's typical operational performance. Active cooling fans were used on both boards during testing, and both ran at close to their rated temperature limits.
- This study also included model optimization using TensorRT [33] inference accelerator tool. It was determined that TensorRT leads to an approximate increase of FPS rate by a factor of 3.5 when compared to the non-optimized smaller variant of yolo-v5 on Nvidia Jetson Nano and Nvidia Jetson Xavier devices.
- After performing model optimization, the Nvidia Jetson produced 11 FPS and Nvidia Jetson Xavier achieved 60 FPS on test data.

## CONCLUSION

Thermal imaging provides superior and effective results in challenging environments such that in low lighting scenarios and has aggregate immunity to visual limitations thus making it an optimal solution for intelligent and safer vehicular systems**.** In this study, we presented a new benchmark thermal dataset that comprises over 35K distinct frames recorded, analyzed, and open-sourced in challenging weather and environmental conditions utilizing a low-cost yet reliable uncooled LWIR thermal camera. All the YOLO v5 network variants were trained using locally gathered data as well as four different publicly available datasets. The performance of trained networks is analysed on both GPU as well as ARM processor-based edge devices for onboard automotive sensor suite feasibility testing. On edge devices, the small and medium network edition of YOLO is deployed and tested due to certain memory limitations and less computational power of these boards. Lastly, we further optimized the smaller network variant using TensorRT inference accelerator to explicitly increase the FPS on edge devices. This allowed the system to achieve 11 frames per second on jetson nano, while the Nvidia Jetson Xavier delivered a significantly higher performance of 60 frames per second. These results validate the potential for thermal imaging as a core component of ADAS systems for intelligent vehicles.

As the future directions, the system's performance can be further enhanced by porting the trained networks on more advanced and powerful edge devices thus tailoring it for real-time onboard deployments. Moreover, the current system focuses on object recognition, but it can be enhanced to incorporate image segmentation, road and lane detection, traffic signal and road signs classification, and object tracking for providing comprehensive driver assistance.

## ACKNOWLEDGMENT

The authors would like to acknowledge Cosmin Rotariu from Xperi-Ireland and the rest of the team members for providing the support in preparing the data accusation setup and helping throughout in data collection and Quentin Noir from Lynred France for giving their feedback. Moreover, the authors would like to acknowledge the contributors of all the public datasets for providing the image resources to carry out this research work and ultralytics for sharing the YOLO-V5 Pytorch version.

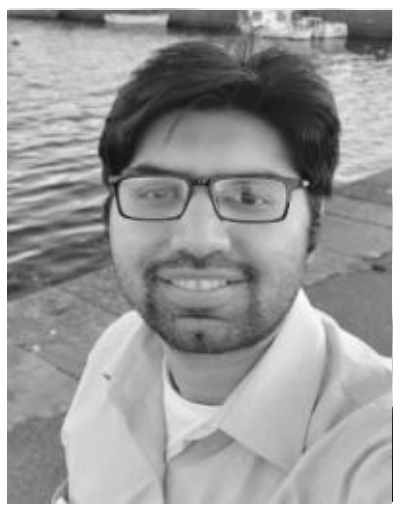

**Muhammad Ali Farooq** received his BE degree in electronic engineering from IQRA University in 2012 and his MS degree in electrical control engineering from the National University of Sciences and Technology (NUST) in 2017. He is currently pursuing the Ph.D. degree at the National University of Ireland Galway (NUIG). His research interests include machine vision, computer vision, video analytics, and sensor fusion. He has won the prestigious H2020 European Union (EU) scholarship and currently working at NUIG as one of the consortium partners in the Heliaus (thermal vision augmented awareness) project funded by EU.

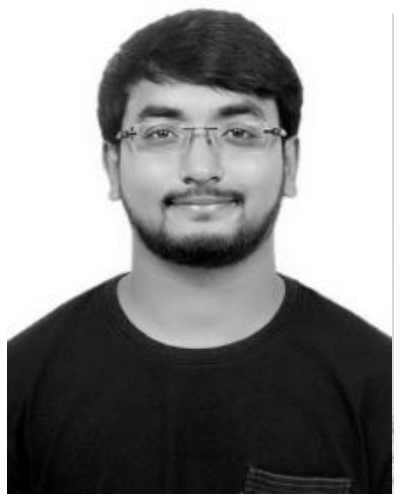

**Waseem Shariff** received his B.E degree in computer science from Nagarjuna College of Engineering and Technology (NCET) in 2019 and his M.S. degree in computer science, specializing in artificial intelligence from National University of Ireland Galway (NUIG) in 2020. He is working as research assistant at National University of Ireland Galway (NUIG). He is associated with Heliaus (thermal vision augmented awareness) project. He is also allied with FotoNation/Xperi research team. His research interests include machine learning utilizing deep neural networks for computer vision applications, including working with synthetic data, thermal data, and RGB.

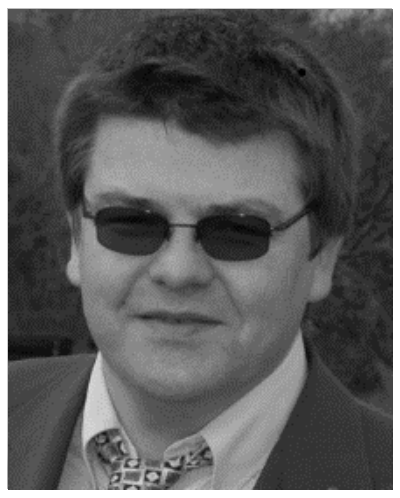

**Peter Corcoran** (Fellow, IEEE) holds a Personal Chair in Electronic Engineering at the College of Science and Engineering, National University of Ireland Galway (NUIG). He was the Co-Founder in several start-up companies, notably FotoNation, now the Imaging Division of Xperi Corporation. He has more than 600 cited technical publications and patents, more than 120 peer-reviewed journal articles, 160 international conference papers, and a co-inventor on more than 300 granted U.S. patents. He is an IEEE Fellow recognized for his contributions to digital camera technologies, notably in-camera red-eye correction and facial detection. He is a member of the IEEE Consumer Technology Society for more than 25 years and the Founding Editor of IEEE Consumer Electronics Magazine.